\documentclass[sigconf,nonacm]{acmart}

\AtBeginDocument{
  }

\usepackage{pgfplots}
\pgfplotsset{compat=1.18}
\usepgfplotslibrary{groupplots}

\usepackage{listings}
\usepackage{upquote}
\usepackage{multirow}
\usepackage{dblfloatfix}
\title{Labeling Training Data for Entity Matching Using Large Language Models}

\author{Aaron Steiner}
\orcid{0009-0006-6946-7057}
\affiliation{
  \institution{Data and Web Science Group\\University of Mannheim}
  \city{Mannheim}
  \country{Germany}}
\email{aaron.steiner@uni-mannheim.de}

\author{Christian Bizer}
\orcid{0000-0003-2367-0237}
\affiliation{
  \institution{Data and Web Science Group\\University of Mannheim}
  \city{Mannheim}
  \country{Germany}}
\email{christian.bizer@uni-mannheim.de}

\renewcommand{\shortauthors}{Steiner and Bizer}

\begin{abstract}
Large language models (LLMs) achieve strong performance on entity matching tasks without requiring task-specific training data, but applying them to large sets of candidate pairs is slow and costly. Matchers built on pretrained language models (PLMs), such as BERT, offer substantially faster inference, but require task-specific training data. This paper presents a systematic experimental study of knowledge-distillation workflows for entity matching in which an LLM teacher labels training pairs and a smaller student matcher is trained on the machine-labeled set. We vary the pair selection strategy, labeling budget, teacher model, correspondence post-processing method, and student model. We evaluate these workflows on eight entity matching benchmarks, including benchmarks that require matching unseen entities and non-English data. We compare students trained on machine-labeled data with matchers trained on the original benchmark training sets. Our experiments show that, in most cases, PLM-based matchers trained on LLM-labeled data perform similarly to those trained on the original benchmark training sets. The pair selection strategy matters most for small labeling budgets, where active learning is often most effective. An open-weight teacher also trains competitive students, so distillation requires no closed-weight models. Compact PLM-based students are able to compete with much larger LLM students on most tasks, while requiring 34 to 459 times less inference time than direct LLM matching. On the two benchmarks with high shares of unseen products, PLM-based students substantially underperform their teachers, as do students trained on the original benchmark data. Under GPT-5.2 pricing, LLM labeling costs per training set are on average \$5.86 to \$8.11. These findings support knowledge distillation as a practical approach to reduce the effort of labeling task-specific training data while enabling efficient inference.

\end{abstract}

\usepackage{needspace}
\begin{document}

\maketitle

\pagestyle{plain}
\noindent\textbf{Artifact Availability:} The source code, data, and other artifacts are available at \url{https://github.com/wbsg-uni-mannheim/Automatic-data-labeling}.

\section{Introduction}

Entity matching (EM) is a core step in data integration. Given one or more collections of records, EM decides which records refer to the same real-world entity \cite{christophides2020overview,christophides2015webdata}. Modern EM systems often use supervised learning, including transformer-based matchers \cite{mudgal2018deep,brunner2020transformer,li2020ditto,barlaug2021survey}. A major limitation of these matchers is their dependence on use case-specific training data in order to achieve high F1 performance \cite{peeters2024wdcproducts}.
Creating such data is costly because useful training sets must include obvious matches and non-matches as well as corner cases: same-family products, compatible components, near-duplicate publication titles, and other pairs that are easy to confuse \cite{christen2012data,peeters2024wdcproducts}. A corner case is a pair whose surface form gives plausible evidence for both labels. Resolving these pairs usually requires human inspection, which raises costs and limits scalability.

Large language models (LLMs) can classify entity pairs directly in a zero-shot setting with strong matching performance~\cite{peeters2025llmem}. At scale, direct LLM matching is slow and expensive, because even after blocking removes obvious non-matches~\cite{christen2012data}, the LLM must be queried for every remaining candidate pair. Pretrained language models (PLMs) such as RoBERTa process high-volume candidate sets faster and with fewer resources, but their accuracy depends on their training set. Teacher/student knowledge distillation~\cite{xu2024survey,10.1145/3699518,fang2026knowledge} connects the two: an LLM teacher labels selected candidate pairs, and a cheaper student matcher is trained on the resulting machine-labeled training set. DistillER~\cite{zeakis2026distiller} studies such workflows.

\begingroup\setlength{\emergencystretch}{1em}
This paper is a systematic experimental study of knowledge-distillation workflows for EM. It varies their components under a common reference to show which design choices matter and relates the resulting students to direct LLM matching in accuracy and runtime. We judge each complete machine-labeled training set, with its own selected pairs and teacher labels, by its \emph{fitness for use} \cite{FelixDataQualityandML2025}, measured as the test F1 of a student against the same student trained on the benchmark training set it replaces.\par\endgroup

Pair selection uses similarity search or one of two active-learning strategies at varying label budgets, and post-processing uses an LLM review method or a graph-based consistency filter. Three teachers assign binary match labels: GPT-5.2, Qwen 3.6 Plus, and the open-weight Kimi K2.6. The students represent three model families: Ditto~\cite{li2020ditto} as a widely used PLM-based matcher, an XGBoost classifier over similarity features, and Qwen3 LLMs with 0.6B, 1.7B, and 8B parameters. The benchmarks combine Abt-Buy, Walmart-Amazon, DBLP-ACM, and DBLP-Scholar from the DeepMatcher suite with benchmarks that each target one challenge: hard non-matches (Dn7 Walmart-Amazon), unseen entities (WDC Products), German-language offers (billiger.de), and structural variations (Semi-HETER).

The findings of this paper are:

\begin{enumerate}
  \setlength{\emergencystretch}{1em}
  \item Machine-labeled training sets are a practical alternative to benchmark training sets for a PLM-based matcher on seven of eight benchmarks.
  \item Compared with its LLM teacher, the compact PLM-based student trained on GPT-5.2 labels comes close to or exceeds it on most benchmarks and competes with much larger LLM students.
  \item On WDC Products and billiger.de, whose test sets contain 100\% and 50\% unseen entities, the student falls behind its teacher with machine or benchmark labels. This confirms the problems of PLM-based matchers with unseen entities reported by Peeters et~al.~\cite{peeters2025llmem} and Wang et~al.~\cite{wang2022bridging}.
  \item The PLM-based student infers more than 30 times faster than direct LLM matching, after a one-time teacher cost of a few dollars in API usage fees.
  \item The pair selection method matters most at small budgets and on hard non-matches, where active learning often needs fewer labels.
  \item The structural profile of a training set, such as its entity coverage, does not predict which set trains the best student.
  \item A human audit of five test sets finds similar error rates for teacher labels and benchmark labels.
\end{enumerate}

Section~\ref{sec:method} presents the knowledge-distillation workflows, Section~\ref{sec:setup} the experimental setup, and Section~\ref{sec:results} the results. Sections~\ref{sec:related}, \ref{sec:limitations}, and~\ref{sec:conclusion} cover related work, limitations, and the conclusion.
\section{The Knowledge Distillation Workflows}
\label{sec:method}

\begin{figure}[h]
  \centering
  \includegraphics[width=\linewidth]{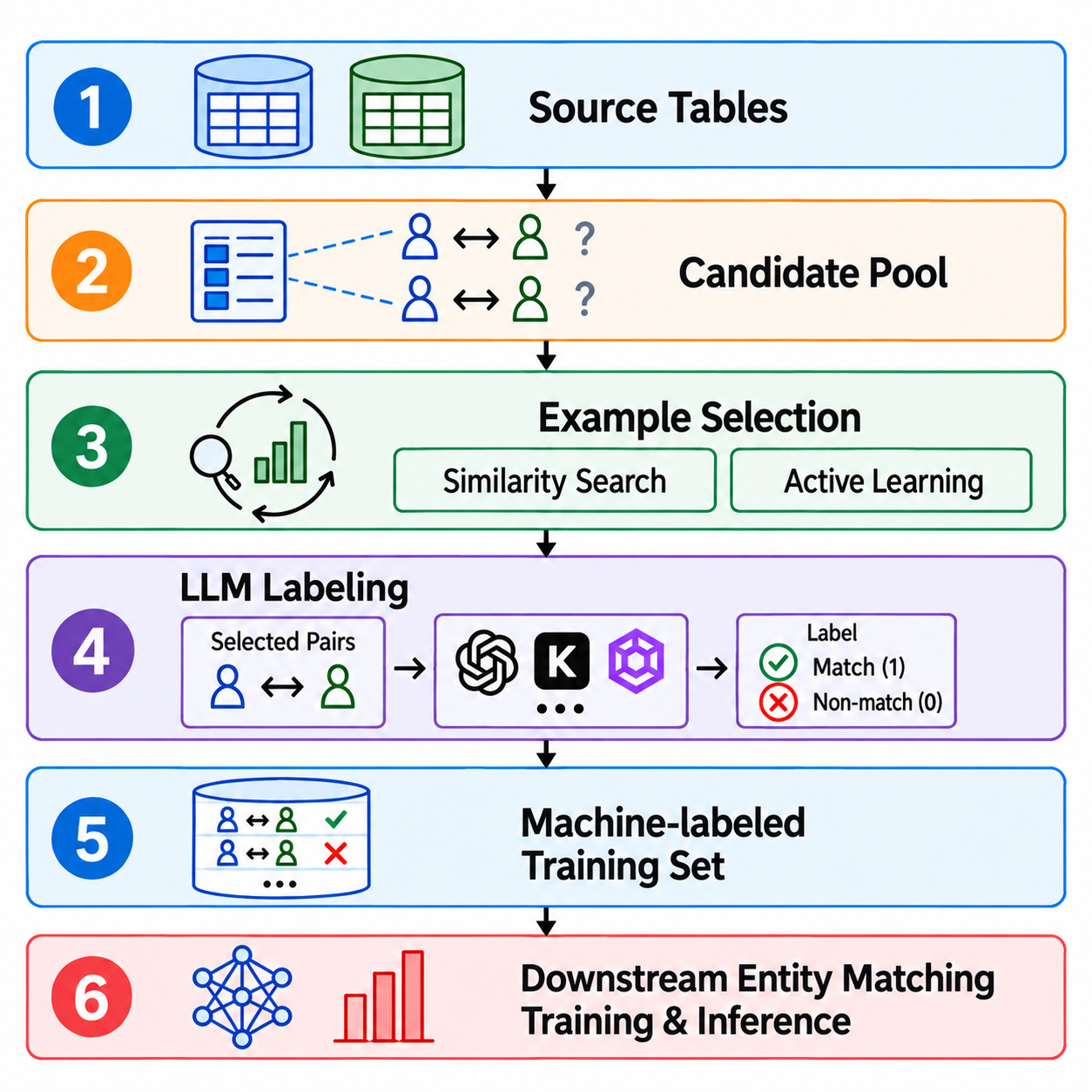}
  \caption{Training-set construction and evaluation workflow.}
  \Description{Workflow diagram showing source tables, candidate-pair construction, selection strategies, LLM labeling, and export of a machine-labeled training set for downstream entity-matching experiments and analysis.}
  \label{fig:workflow}
\end{figure}

This section describes the knowledge distillation workflows that we compare in the paper. Figure~\ref{fig:workflow} illustrates the main steps of the workflows: given two source tables, the workflows build a candidate pool, select pairs for labeling, let an LLM teacher label the pairs, optionally post-process the labels, and export a machine-labeled training set for training a student matcher. The experiments vary four design choices: pair-selection strategy, teacher model, label post-processing method, and student model.

\subsection{Candidate Pool Generation}
\label{sec:candidate-pool}

The workflows start with initializing the candidate pool from which the pair selection methods will later select pairs for labeling.
The benchmark training sets contain record pairs selected by the benchmark authors. To reconstruct the source tables before pair selection, we split each training set into two datasets: One table containing all left-side records and one table containing all right-side records of the benchmark training set. This keeps record selection of the benchmark intact, but undoes the benchmark's pair selection. The workflows thus start from two unlabeled source tables.

The pool uses embedding retrieval. We embed the records with \path{text-embedding-3-small}\footnote{\url{https://platform.openai.com/docs/models/text-embedding-3-small}} and rank right-table records by cosine similarity to each left-table record. For each query, we select 18 nearest neighbors and sample two candidates from the lower-ranked half. The neighbors supply plausible matches and hard non-matches, the lower-ranked samples easier negatives. Duplicate pairs are removed before labeling, so the pool size depends on the distinct pairs.

\subsection{Pair Selection}

The pair-selection strategy determines which candidate pairs the teacher labels. We compare three strategies: similarity search, active learning with a feature-based committee (AL-ML), and active learning with a Ditto committee (AL-Ditto).

\textbf{Similarity search:} For each left-table record, the strategy sorts the record's pool candidates by descending embedding similarity and sends them to the teacher in batches of five. Labeling continues until the labeled candidates contain at least one positive and four negatives, or until the candidates or the API budget are exhausted. If at least two positives have been found, the record lies in a dense positive region, and the teacher labels all its remaining candidates. Otherwise, the teacher also labels the two lowest-ranked candidates not yet processed, which adds easy negatives, and the strategy retains the positives and negatives. A record whose first batch satisfies the stopping condition thus costs seven teacher calls, whereas a first batch without a positive continues the scan. The final training set is drawn from the retained pairs at the target size.

\textbf{Active learning:} Both active-learning strategies start from a similarity-search seed that targets 100 pairs with a 30/70 positive/negative split. This fixed ratio gives the first committee both classes despite the negative-dominated pool. In each round, the strategy trains a committee of matching models on the currently labeled pairs, scores up to 20{,}000 unlabeled pool pairs, and passes the pairs on which the committee disagrees most to the teacher.

\textbf{AL-ML:} This variant uses a committee of logistic regression, random forests, extra trees, gradient-boosted trees, and rule-based matchers. Each pair is represented by one similarity score per attribute plus the cosine similarity of the two text-embedding-3-small record embeddings. Text attributes such as titles use token Jaccard similarity, and numeric attributes such as price or year use normalized closeness, $1 - |a-b| / \max(|a|, |b|)$. The committee ranks candidates by disagreement over match-probability scores, and the workflow labels up to 500 new pairs per round.

\textbf{AL-Ditto:} This variant uses the same seed and candidate pool but trains a bagged ensemble of five Ditto matchers, each on a bootstrap sample of the labeled pairs with a different seed. The ensemble scores up to 20{,}000 unlabeled candidates, and the teacher labels the 500 pairs on which it disagrees most. Ranking is lexicographic: variance of the match probabilities first, then entropy of the thresholded match/non-match votes, distance of the mean probability from 0.5, and number of split votes. New labels join the training set, and the loop repeats until the target size is reached, the budget is spent, the pool is empty, or a round adds no pairs.

In active-learning terms, both committees perform uncertainty sampling by ranking candidates by disagreement, with two forms of ensembling: a heterogeneous feature-based committee (AL-ML) and a bagged Ditto committee (AL-Ditto). Similarity search is a ranking heuristic that concentrates labels where matches are most likely. All three strategies draw from the retrieval pool, which favors plausible matches and hard non-matches (Section~\ref{sec:candidate-pool}). Diversity-based selection, density-weighted uncertainty, and budget-aware methods~\cite{genossar2023battleship,pulsone2026beacon,maciejewski2025progressive} remain outside the comparison.

\subsection{LLM-based Labeling}

For every selected pair, the teacher receives the two records serialized as JSON and returns \texttt{\{"match": true|false\}} (Listing~\ref{lst:prompt}). We adopt the \emph{general-complex} task description of \citet{peeters2025llmem} because it uses domain-independent wording and applies unchanged to product and bibliographic records, and add the JSON-only constraint for reliable parsing. If an output cannot be parsed, the retry instruction is appended to the user message, with up to four attempts per pair. Most experiments use GPT-5.2\footnote{\url{https://platform.openai.com/docs/models/gpt-5.2}} as the teacher model. The teacher comparison in Section~\ref{sec:labeler} also employs Qwen 3.6 Plus\footnote{\url{https://openrouter.ai/qwen/qwen3.6-plus}} and the open-weight Kimi K2.6\footnote{\url{https://openrouter.ai/moonshotai/kimi-k2.6}}.

\begin{lstlisting}[caption={Teacher prompt for labeling one record pair.},label={lst:prompt}]
System: You are an expert entity matcher. Decide if two records refer to the same real-world entity. Return only valid JSON with exactly one field: {"match": true|false}.

User: Do the two entity descriptions refer to the same real-world entity? Entity 1: '{left_json}'. Entity 2: '{right_json}'.

Retry, appended to the user message after an invalid output: Your previous output was invalid. Return valid JSON with exactly one field: {"match": true|false}.
\end{lstlisting}

\subsection{Post-Processing the Labels}
Post-processing corrects wrong teacher labels or drops pairs whose labels are likely wrong. We study two methods and their combinations: a second-model review pass over each label and a graph-based consistency check across the labeled set.

\textbf{Relabeling:} Our initial experiments showed that uncertain cases often receive positive labels. The relabeling method labels every pair again with GPT-5-mini, using a conservative review prompt that treats conflicting model numbers, editions, capacities, or variants as evidence against a match and does not show the assigned label. The repository contains the prompt. It returns a decision and a confidence score. In the relabel variant, the reviewed decision replaces the original decision. In the relabel-drop variant, a changed decision marks the original label as unstable and the pair is dropped. The confidence score proved too noisy for filtering and is not used.

\textbf{Closure-based dropping:} This filter enforces structural consistency. We build an undirected graph whose nodes are source records and whose edges are the pairs labeled as matches. First, in components of at least three records, the filter drops bridge edges, match pairs whose removal would split the component, because no second positive path supports that merge. Two-record components keep their single edge. Second, it drops negative pairs whose records remain connected by a positive path in the retained graph, because such a negative contradicts the transitive closure of the matches. The retained labels contain no negative pair inside a positive component. This is structural consistency: correct pairs can be removed along with erroneous ones.

The two filters can be combined in two ways. The union variant drops any pair flagged by either filter, the strictest setting we report, while the intersection variant drops only pairs flagged by both.
\section{Experimental Setup}
\label{sec:setup}

We compare matchers trained on benchmark training sets with matchers trained on the machine-labeled sets from Section~\ref{sec:method}. All comparisons use the benchmark test splits, a fixed student configuration, and one evaluation script. The main comparison uses GPT-5.2 as teacher, Ditto as student, and no post-processing. We then vary label budgets, teachers, post-processing, and students.

For each fixed training set, F1 is reported in percentage points as mean and sample standard deviation over three or more student training runs with different seeds. These repetitions measure student-training variability conditional on one acquired set. Welch's t-tests compare the student F1 for each machine-labeled set with that of benchmark training, Holm-corrected over the three strategies per benchmark ($\alpha = 0.05$). As non-significance does not imply equivalence, we also report observed gaps. Zero-shot results use one run. The artifact contains run-level results. Bold marks the best F1 per benchmark, underlining the second best. In Tables~\ref{tab:selection-strategies}, \ref{tab:llm-comparison}, and~\ref{tab:ditto-refinements}, $\Delta$ is the best machine-labeled F1 minus benchmark-trained F1.

We use eight benchmarks from two domains, products and literature (Table~\ref{tab:dataset-statistics}). Product matching involves dealing with product variants, accessories, and kits, bibliographic matching involves title abbreviations, venue acronyms, and author-name variants. They combine the DeepMatcher benchmarks with benchmarks for specific challenges.

\paragraph{DeepMatcher benchmarks.}
\citet{koepcke2010evaluation} compiled Abt-Buy, DBLP-ACM, and DBLP-Scholar. All four, including Walmart-Amazon, belong to the benchmark suite used to evaluate DeepMatcher~\cite{mudgal2018deep} and Ditto~\cite{li2020ditto}. We use their splits for comparability with prior work. \citet{papadakis2024critical} classify the two product tasks as challenging and the two DBLP tasks as easy, which shows how the workflows behave on nearly saturated tasks.
\paragraph{Hard non-matches.}
\citet{papadakis2024critical} find that most established benchmarks pose rather easy classification tasks and rebuild several of them as harder ones. Their Dn7 Walmart-Amazon uses the same records as Walmart-Amazon, but only 1.8\% of its pairs are matches, compared with 9.4\%. The best threshold on token-based cosine or Jaccard similarity reaches an F1 below 0.5, so Dn7 tests pair selection when surface similarity is a weak signal.

\paragraph{Unseen entities.}
WDC Products~\cite{peeters2024wdcproducts} varies the share of corner cases and the share of test entities unseen during training. We build the source tables from its large training split with 80\% corner cases and evaluate on the test set with 100\% unseen entities. These entities appear neither in the source tables nor in the benchmark training set, so the benchmark tests how well a trained matcher generalizes to new products.

\paragraph{Broad product range.}
billiger.de~\cite{billiger2026benchmark} adopts the design of WDC Products for German offers from the price comparison platform billiger.de. It covers thirteen product categories, including clothing and furniture. Its corner cases include closely related product variants. We build the source tables from its large training split with 80\% corner cases, use its medium training split with 5,897 pairs as benchmark training set, and evaluate on the recommended default test set with 50\% unseen products. It thus combines unseen entities, a broad product range, and German-language records. Its labels and splits were first published in August 2026, after GPT-5.2's documented training cutoff of 31 August 2025,\footnote{\url{https://developers.openai.com/api/docs/models/gpt-5.2}} and after the April 2026 releases of Qwen 3.6 Plus and Kimi K2.6, and thus fall outside the pretraining windows of all three teachers. The underlying products may have been known from the web.

\paragraph{Structural Variation.}
Machamp~\cite{wang2021machamp} extends EM benchmarks from structured tables with a shared schema to semi-structured records of differing structure. Its Semi-HETER task matches two tables of book records with 127 and 56 distinct attribute combinations and has the smallest training set of the eight benchmarks. \citet{zhang2026beyondscale} retain Semi-HETER among the Machamp tasks on which existing methods have not reached near-saturated performance. As schema matching would in an integration pipeline, we map the attribute names to shared attributes such as title, authors, and publisher. After this mapping, the records still differ in their filled attributes and value formats: author lists, for example, are separated by semicolons or commas and partly carry roles such as editor.

\begin{table}[t]
\centering
\caption{Benchmark sizes: distinct records and labeled pairs per split.}
\label{tab:dataset-statistics}
\small
\setlength{\tabcolsep}{2.5pt}
\begin{tabular}{lrrrr}
\toprule
Dataset & Records (L/R) & Train & Valid. & Test \\
\midrule
    \multicolumn{5}{l}{\textit{Products}} \\
Abt-Buy & 1,025 / 1,009 & 5,743 & 1,916 & 1,916 \\
Walmart-Amazon & 1,579 / 4,533 & 6,144 & 2,049 & 2,049 \\
Dn7 Walmart-Amazon & 2,554 / 10,717 & 26,050 & 8,684 & 8,684 \\
WDC Products (100\% unseen) & 2,838 / 2,840 & 19,835 & 3,500 & 4,500 \\
billiger.de (50\% unseen) & 3,494 / 3,495 & 5,897 & 4,452 & 4,437 \\
    \midrule
    \multicolumn{5}{l}{\textit{Literature}} \\
DBLP-ACM & 2,306 / 2,110 & 7,417 & 2,473 & 2,473 \\
DBLP-Scholar & 2,504 / 9,494 & 17,223 & 5,742 & 5,742 \\
Semi-HETER & 994 / 1,049 & 1,240 & 414 & 414 \\
\bottomrule
\end{tabular}
\end{table}

\paragraph{Teachers.}
GPT-5.2 serves as the main teacher. Qwen 3.6 Plus adds a second proprietary provider, and the open-weight Kimi K2.6 can run locally on sufficient hardware, which matters when sensitive records cannot leave the organization.

\paragraph{Students.}
The students represent three model families. Ditto~\cite{li2020ditto} fine-tunes RoBERTa-base and is a PLM-based matcher that recent benchmark studies use as a reference~\cite{peeters2024wdcproducts,papadakis2024critical,wang2021machamp}. XGBoost over the similarity features of the AL-ML committee (Section~\ref{sec:method}) represents classical feature-based matchers. Qwen3 offers small LLMs in several sizes, and its 0.6B\footnote{\url{https://huggingface.co/Qwen/Qwen3-0.6B}}, 1.7B\footnote{\url{https://huggingface.co/Qwen/Qwen3-1.7B}}, and 8B\footnote{\url{https://huggingface.co/Qwen/Qwen3-8B}} models, fine-tuned with LoRA as in~\citet{steiner2025finetuning}, show how student size affects accuracy. Each student family uses one fixed configuration, and the repository holds all hyperparameters.
\section{Results}
\label{sec:results}

This section reports the experiments along the design choices of the workflows and relates their outcomes to cost and training-set quality. Section~\ref{sec:f1-comparison} compares the fitness for use of machine-labeled and benchmark training sets under the three pair-selection strategies. Sections~\ref{sec:budget} to~\ref{sec:student-models} then vary one design choice at a time: the label budget, the teacher model, label post-processing, and the student model. Section~\ref{sec:direct-llm-matching} compares the teachers as direct matchers with the trained students, and Section~\ref{sec:cost} relates construction cost to inference runtime. Sections~\ref{sec:composition} and~\ref{sec:error-analysis} analyze how the selection strategies shape the training sets and how often teachers and benchmarks mislabel pairs. Each subsection ends with a short finding that states what the experiment shows and what it implies for practitioners who build training sets with LLMs. Where results differ by benchmark challenge (Section~\ref{sec:setup}), the finding names it.

\subsection{Machine-labeled Versus Benchmark Training Sets}
\label{sec:f1-comparison}

We compare the fitness for use of machine-labeled and benchmark training sets, measured by the F1 of a Ditto student on the benchmark test set (Table~\ref{tab:selection-strategies}). The machine-labeled sets come from the three pair-selection strategies with GPT-5.2 as teacher. The sets differ from the benchmark sets in both the selected pairs and their labels, so the comparison measures whole training set quality.

\begin{table}[!t]
  \centering
  \caption{Ditto F1 by training source (GPT-5.2 teacher). $\uparrow$/$\downarrow$: student F1 significantly above/below benchmark training.}
  \label{tab:selection-strategies}
  \label{tab:extended}
  \footnotesize
  \setlength{\tabcolsep}{1pt}
  \renewcommand{\arraystretch}{1.12}
  \newcommand{\tabletworesult}[2]{#1\,{\scriptsize $\pm$\,#2}}
  \newcommand{\tabletwoname}[1]{\begin{tabular}[c]{@{}l@{}}#1\end{tabular}}
  \begin{tabular*}{\columnwidth}{@{\extracolsep{\fill}}lrrrrr@{}}
    \toprule
    Benchmark & \shortstack{Benchmark\\training set} & \shortstack{Sim.\\search} & AL-ML & AL-Ditto & $\Delta$ \\
    \midrule
    \multicolumn{6}{l}{\textit{Products}} \\
    \textsc{Abt-Buy} & \tabletworesult{\underline{88.34}}{1.54} & \tabletworesult{86.77}{0.09} & \tabletworesult{87.76}{0.67} & \tabletworesult{\textbf{89.02}}{0.11} & $+0.68$ \\
        \tabletwoname{\textsc{Walmart-}\\\textsc{Amazon}} & \tabletworesult{85.66}{1.04} & \tabletworesult{85.37}{1.55} & \tabletworesult{\underline{86.88}}{0.43} & \tabletworesult{\textbf{87.25}}{0.65} & $+1.59$ \\
    \tabletwoname{Dn7 Walmart-\\Amazon} & \tabletworesult{72.20}{3.93} & \tabletworesult{70.60}{0.97} & \tabletworesult{\textbf{76.22}}{1.22} & \tabletworesult{\underline{75.64}}{2.86} & $+4.02$ \\
        \tabletwoname{\textsc{WDC Products}} & \tabletworesult{\underline{71.23}}{0.83} & \tabletworesult{69.53}{1.25} & \tabletworesult{70.65}{0.42} & \tabletworesult{\textbf{72.17}}{1.01} & $+0.94$ \\
    \tabletwoname{billiger.de} & \tabletworesult{\textbf{65.09}}{2.11} & \tabletworesult{58.85}{2.70} & \tabletworesult{56.28$^{\downarrow}$}{1.36} & \tabletworesult{\underline{60.00}}{1.31} & $-5.09$ \\
        \midrule
    \multicolumn{6}{l}{\textit{Literature}} \\
    \textsc{DBLP-ACM} & \tabletworesult{\textbf{98.43}}{0.34} & \tabletworesult{\underline{97.99}}{0.40} & \tabletworesult{97.50}{0.17} & \tabletworesult{97.91}{0.60} & $-0.44$ \\
        \tabletwoname{\textsc{DBLP-}\\\textsc{Scholar}} & \tabletworesult{\textbf{95.64}}{0.26} & \tabletworesult{93.54$^{\downarrow}$}{0.18} & \tabletworesult{93.81$^{\downarrow}$}{0.38} & \tabletworesult{\underline{93.86}$^{\downarrow}$}{0.23} & $-1.79$ \\
        Semi-HETER & \tabletworesult{82.97}{5.29} & \tabletworesult{91.33$^{\uparrow}$}{1.92} & \tabletworesult{\textbf{93.13}$^{\uparrow}$}{3.62} & \tabletworesult{\underline{91.53}$^{\uparrow}$}{2.24} & $+10.15$ \\
        \bottomrule
  \end{tabular*}
\end{table}

\begin{table}[!t]
  \centering
  \caption{Ditto F1 on benchmark training pairs by label source. $\downarrow$: significantly below original. $\Delta$: GPT-5.2 minus original.}
  \label{tab:fixed-pairs}
  \footnotesize
  \setlength{\tabcolsep}{3pt}
  \renewcommand{\arraystretch}{1.08}
  \begin{tabular*}{\columnwidth}{@{\extracolsep{\fill}}lrrr@{}}
    \toprule
    Benchmark & Original labels & GPT-5.2 labels & $\Delta$ \\
    \midrule
    \multicolumn{4}{l}{\textit{Products}} \\
    \textsc{Abt-Buy} & 88.34 $\pm$ 1.54 & 86.57 $\pm$ 0.40 & $-1.77$ \\
    \textsc{Walmart-Amazon} & 85.66 $\pm$ 1.04 & 86.25 $\pm$ 1.83 & $+0.59$ \\
    Dn7 Walmart-Amazon & 72.20 $\pm$ 3.93 & 72.73 $\pm$ 0.97 & $+0.53$ \\
    \textsc{WDC Products} & 71.23 $\pm$ 0.83 & 70.90 $\pm$ 1.22 & $-0.33$ \\
    billiger.de & 65.09 $\pm$ 2.11 & 62.64 $\pm$ 1.11 & $-2.45$ \\
    \midrule
    \multicolumn{4}{l}{\textit{Literature}} \\
    \textsc{DBLP-ACM} & 98.43 $\pm$ 0.34 & 97.83 $\pm$ 0.18 & $-0.60$ \\
    \textsc{DBLP-Scholar} & 95.64 $\pm$ 0.26 & 93.26$^{\downarrow}$ $\pm$ 0.48 & $-2.38$ \\
    Semi-HETER & 82.97 $\pm$ 5.29 & 83.74 $\pm$ 6.37 & $+0.77$ \\
    \bottomrule
  \end{tabular*}
\end{table}

\begin{figure*}[!t]
  \centering
  \begin{tikzpicture}
    \begin{groupplot}[
      group style={group name=budgetplots, group size=4 by 2, horizontal sep=0.65cm, vertical sep=0.85cm},
      scale only axis,
      width=0.205\linewidth,
      height=2.6cm,
      xlabel={},
      scaled x ticks=base 10:-3,
      xtick scale label code/.code={},
      ylabel={},
      tick label style={font=\footnotesize},
      label style={font=\small},
      title style={font=\footnotesize},
      every axis plot/.append style={mark size=1.1pt},
      legend columns=4,
      legend style={draw=none, font=\footnotesize, /tikz/every even column/.append style={column sep=7pt}},
      grid=major,
      grid style={gray!20},
      cycle list name=color list,
    ]

    \nextgroupplot[legend to name=budgetlegend, legend entries={AL-Ditto, AL-ML, Similarity search, Benchmark set}, title=\textsc{Abt-Buy}, xmin=0, xmax=6500, ymin=72, ymax=93, ytick={75,80,85,90}]
    \addplot[mark=*, thick, blue!70!black, error bars/.cd, y dir=both, y explicit] coordinates {
      (1000,78.66) +- (0,1.09)
      (2000,88.06) +- (0,0.59)
      (3000,89.49) +- (0,0.83)
      (4000,89.41) +- (0,0.77)
      (5000,88.18) +- (0,0.93)
      (6000,89.02) +- (0,0.11)
    };
    \addplot[mark=square*, thick, orange!85!black, error bars/.cd, y dir=both, y explicit] coordinates {
      (1000,80.48) +- (0,1.60)
      (2000,84.26) +- (0,1.55)
      (3000,86.64) +- (0,1.00)
      (4000,86.87) +- (0,0.57)
      (5000,87.07) +- (0,0.51)
      (6000,87.76) +- (0,0.67)
    };
    \addplot[mark=triangle*, thick, gray!55!black, error bars/.cd, y dir=both, y explicit] coordinates {
      (1000,77.55) +- (0,0.43)
      (2000,81.86) +- (0,1.88)
      (3000,83.76) +- (0,2.39)
      (4000,86.67) +- (0,1.00)
      (5000,86.05) +- (0,0.29)
      (6000,86.77) +- (0,0.09)
    };
    \addplot[dashed, thick, black, no markers] coordinates {(0,88.34) (6500,88.34)};

    \nextgroupplot[title=\textsc{Walmart-Amazon}, xmin=0, xmax=6500, ymin=71.03, ymax=91.03, ytick={75,80,85,90}]
    \addplot[mark=*, thick, blue!70!black, error bars/.cd, y dir=both, y explicit] coordinates {
      (1000,84.51) +- (0,1.47)
      (2000,86.39) +- (0,0.69)
      (3000,86.51) +- (0,0.76)
      (4000,86.77) +- (0,1.15)
      (5000,85.88) +- (0,1.22)
      (6000,86.73) +- (0,1.72)
    };
    \addplot[mark=square*, thick, orange!85!black, error bars/.cd, y dir=both, y explicit] coordinates {
      (1000,82.52) +- (0,1.54)
      (2000,84.79) +- (0,2.48)
      (3000,86.98) +- (0,0.43)
      (4000,86.82) +- (0,0.43)
      (5000,86.40) +- (0,1.01)
      (6000,86.89) +- (0,1.35)
    };
    \addplot[mark=triangle*, thick, gray!55!black, error bars/.cd, y dir=both, y explicit] coordinates {
      (1000,79.54) +- (0,4.06)
      (2000,84.74) +- (0,1.46)
      (3000,84.22) +- (0,1.11)
      (4000,87.31) +- (0,0.66)
      (5000,83.75) +- (0,0.17)
      (6000,86.07) +- (0,2.04)
    };
    \addplot[dashed, thick, black, no markers] coordinates {(0,85.66) (6500,85.66)};

    \nextgroupplot[title=\textsc{WDC Products}, xmin=0, xmax=20000, ymin=50, ymax=80, ytick={50,60,70,80}, xtick={0,5000,10000,15000,20000}]
    \addplot[mark=*, thick, blue!70!black, error bars/.cd, y dir=both, y explicit] coordinates {
      (1000,51.62) +- (0,4.53)
      (2000,64.20) +- (0,2.96)
      (3000,66.56) +- (0,2.25)
      (4000,67.19) +- (0,0.75)
      (5000,68.10) +- (0,0.95)
      (6000,68.61) +- (0,1.06)
      (7000,70.32) +- (0,1.71)
      (8000,70.60) +- (0,0.84)
      (9000,69.83) +- (0,1.36)
      (10000,71.08) +- (0,0.70)
      (11000,70.47) +- (0,2.04)
      (12000,70.04) +- (0,1.07)
      (13000,71.76) +- (0,0.34)
      (14000,71.63) +- (0,1.28)
      (15000,72.43) +- (0,0.77)
      (16000,71.24) +- (0,0.77)
      (17000,72.55) +- (0,1.36)
      (18000,72.23) +- (0,1.04)
      (19000,72.01) +- (0,1.40)
    };
    \addplot[mark=square*, thick, orange!85!black, error bars/.cd, y dir=both, y explicit] coordinates {
      (1000,52.67) +- (0,1.76)
      (2000,62.00) +- (0,2.87)
      (3000,65.76) +- (0,2.03)
      (4000,66.93) +- (0,2.41)
      (5000,67.07) +- (0,0.90)
      (6000,68.88) +- (0,1.37)
      (7000,69.13) +- (0,1.15)
      (8000,68.44) +- (0,2.07)
      (9000,69.44) +- (0,1.42)
      (10000,68.60) +- (0,2.33)
      (11000,70.87) +- (0,0.52)
      (12000,70.79) +- (0,0.58)
      (13000,70.48) +- (0,0.39)
      (14000,70.15) +- (0,1.28)
      (15000,71.15) +- (0,1.01)
      (16000,68.84) +- (0,0.55)
      (17000,69.85) +- (0,1.34)
      (18000,70.96) +- (0,0.96)
      (19000,70.56) +- (0,1.09)
    };
    \addplot[mark=triangle*, thick, gray!55!black, error bars/.cd, y dir=both, y explicit] coordinates {
      (1000,58.77) +- (0,0.69)
      (2000,63.03) +- (0,2.54)
      (3000,65.11) +- (0,0.29)
      (4000,66.01) +- (0,0.85)
      (5000,67.03) +- (0,1.71)
      (6000,67.61) +- (0,2.44)
      (7000,69.37) +- (0,1.39)
      (8000,69.15) +- (0,1.21)
      (9000,69.75) +- (0,1.84)
      (10000,68.44) +- (0,0.61)
      (11000,69.10) +- (0,0.61)
      (12000,68.66) +- (0,0.93)
      (13000,68.71) +- (0,1.39)
      (14000,69.03) +- (0,0.88)
      (15000,69.08) +- (0,1.85)
      (16000,69.91) +- (0,1.27)
      (17000,69.93) +- (0,1.40)
      (18000,70.17) +- (0,1.72)
      (19000,69.37) +- (0,0.49)
    };
    \addplot[dashed, thick, black, no markers] coordinates {(0,71.23) (20000,71.23)};

    \nextgroupplot[title=\textsc{DBLP-ACM}, xmin=0, xmax=7500, ymin=94.19, ymax=99.99, ytick={95,97,99}]
    \addplot[mark=*, thick, blue!70!black, error bars/.cd, y dir=both, y explicit] coordinates {
      (1000,97.65) +- (0,0.38)
      (2000,98.14) +- (0,0.34)
      (3000,97.47) +- (0,0.45)
      (4000,97.48) +- (0,0.48)
      (5000,97.60) +- (0,0.18)
      (6000,97.69) +- (0,0.28)
      (7000,97.69) +- (0,0.30)
    };
    \addplot[mark=square*, thick, orange!85!black, error bars/.cd, y dir=both, y explicit] coordinates {
      (1000,97.40) +- (0,0.15)
      (2000,97.83) +- (0,0.44)
      (3000,97.29) +- (0,0.47)
      (4000,97.50) +- (0,0.34)
      (5000,97.69) +- (0,0.61)
      (6000,97.58) +- (0,0.22)
      (7000,97.72) +- (0,0.58)
    };
    \addplot[mark=triangle*, thick, gray!55!black, error bars/.cd, y dir=both, y explicit] coordinates {
      (1000,97.45) +- (0,0.46)
      (2000,97.83) +- (0,0.26)
      (3000,97.68) +- (0,0.56)
      (4000,97.75) +- (0,0.29)
      (5000,98.03) +- (0,0.64)
      (6000,97.77) +- (0,0.22)
      (7000,97.63) +- (0,0.41)
    };
    \addplot[dashed, thick, black, no markers] coordinates {(0,98.43) (7500,98.43)};

    \nextgroupplot[title=\textsc{DBLP-Scholar}, xmin=0, xmax=17500, ymin=83.94, ymax=99.94, ytick={85,90,95}, xtick={0,5000,10000,15000}]
    \addplot[mark=*, thick, blue!70!black, error bars/.cd, y dir=both, y explicit] coordinates {
      (1000,93.59) +- (0,0.57)
      (2000,93.38) +- (0,0.33)
      (3000,94.24) +- (0,0.35)
      (4000,93.57) +- (0,0.37)
      (5000,94.28) +- (0,0.15)
      (6000,93.70) +- (0,0.33)
      (7000,93.84) +- (0,0.17)
      (8000,93.95) +- (0,0.10)
      (9000,94.01) +- (0,0.80)
      (10000,93.80) +- (0,0.30)
      (11000,93.30) +- (0,0.61)
      (12000,94.05) +- (0,0.38)
      (13000,94.02) +- (0,0.44)
      (14000,93.72) +- (0,0.39)
      (15000,94.00) +- (0,0.17)
      (16000,93.43) +- (0,0.52)
      (17000,93.84) +- (0,0.24)
    };
    \addplot[mark=square*, thick, orange!85!black, error bars/.cd, y dir=both, y explicit] coordinates {
      (1000,92.57) +- (0,0.48)
      (2000,92.29) +- (0,0.73)
      (3000,92.81) +- (0,0.91)
      (4000,92.89) +- (0,0.49)
      (5000,93.86) +- (0,0.34)
      (6000,92.76) +- (0,0.17)
      (7000,93.50) +- (0,0.27)
      (8000,93.64) +- (0,0.32)
      (9000,93.55) +- (0,0.47)
      (10000,93.58) +- (0,0.19)
      (11000,93.91) +- (0,0.12)
      (12000,93.70) +- (0,0.15)
      (13000,93.78) +- (0,0.51)
      (14000,94.08) +- (0,0.26)
      (15000,93.60) +- (0,0.62)
      (16000,93.89) +- (0,0.26)
      (17000,93.11) +- (0,0.13)
    };
    \addplot[mark=triangle*, thick, gray!55!black, error bars/.cd, y dir=both, y explicit] coordinates {
      (1000,91.74) +- (0,1.51)
      (2000,92.45) +- (0,0.15)
      (3000,92.71) +- (0,0.21)
      (4000,92.59) +- (0,1.01)
      (5000,93.62) +- (0,0.31)
      (6000,93.30) +- (0,0.49)
      (7000,93.25) +- (0,0.43)
      (8000,92.74) +- (0,0.47)
      (9000,93.57) +- (0,0.56)
      (10000,93.36) +- (0,0.49)
      (11000,92.96) +- (0,0.53)
      (12000,93.01) +- (0,0.46)
      (13000,93.24) +- (0,0.32)
      (14000,93.22) +- (0,0.55)
      (15000,93.41) +- (0,0.20)
      (16000,93.78) +- (0,0.30)
      (17000,93.64) +- (0,0.29)
    };
    \addplot[dashed, thick, black, no markers] coordinates {(0,95.64) (17500,95.64)};

    \nextgroupplot[title={\textsc{Semi-HETER}}, xmin=0, xmax=1300, ymin=65, ymax=100, xtick={0,300,600,900,1200}]
    \addplot[mark=*, thick, blue!70!black, mark size=1.4pt, error bars/.cd, y dir=both, y explicit] coordinates {
      (300,80.174743) +- (0,5.084831)
      (600,84.996523) +- (0,8.261151)
      (900,90.490294) +- (0,0.811166)
      (1200,92.467340) +- (0,1.364594)
      (1240,91.530486) +- (0,2.236070)
    };
    \addplot[mark=square*, thick, orange!85!black, mark size=1.4pt, error bars/.cd, y dir=both, y explicit] coordinates {
      (300,84.386430) +- (0,4.159839)
      (600,80.995631) +- (0,11.182530)
      (900,89.260191) +- (0,4.225207)
      (1200,93.520260) +- (0,2.916575)
      (1240,93.126830) +- (0,3.618607)
    };
    \addplot[mark=triangle*, thick, gray!55!black, mark size=1.4pt, error bars/.cd, y dir=both, y explicit] coordinates {
      (300,88.725935) +- (0,5.585777)
      (600,92.731860) +- (0,2.366572)
      (900,91.142579) +- (0,1.893787)
      (1200,91.101983) +- (0,2.130416)
      (1240,91.326071) +- (0,1.922105)
    };
    \addplot[dashed, thick, black, no markers] coordinates {(0,82.972245) (1300,82.972245)};

    \nextgroupplot[title={\textsc{Dn7 Walmart-Amazon}}, xmin=0, xmax=27000, ymin=35, ymax=80, xtick={0,5000,10000,15000,20000,25000}]
    \addplot[mark=*, thick, blue!70!black, mark size=1.4pt, error bars/.cd, y dir=both, y explicit] coordinates {
      (1000,69.013302) +- (0,1.075718)
      (2000,74.419070) +- (0,1.833654)
      (3000,74.548315) +- (0,1.602597)
      (4000,74.791306) +- (0,0.557632)
      (5000,75.868305) +- (0,0.104257)
      (6000,75.305005) +- (0,0.429699)
      (7000,75.869795) +- (0,1.082439)
      (8000,75.067966) +- (0,1.283185)
      (9000,76.172604) +- (0,1.047916)
      (10000,74.590627) +- (0,1.677883)
      (11000,74.193490) +- (0,2.215539)
      (12000,74.487502) +- (0,2.490596)
      (13000,74.100790) +- (0,2.814241)
      (14000,74.781660) +- (0,1.045179)
      (15000,73.909907) +- (0,1.177236)
      (16000,75.938111) +- (0,0.937486)
      (17000,74.470400) +- (0,3.380981)
      (18000,74.735512) +- (0,0.927873)
      (19000,74.510117) +- (0,1.195004)
      (20000,75.511492) +- (0,0.960633)
      (21000,76.185626) +- (0,0.203561)
      (22000,75.321576) +- (0,0.035284)
      (23000,75.217512) +- (0,1.310344)
      (24000,76.334588) +- (0,1.022409)
      (25000,74.962219) +- (0,0.909181)
      (26000,75.071939) +- (0,1.713369)
      (26050,75.644241) +- (0,2.859474)
    };
    \addplot[mark=square*, thick, orange!85!black, mark size=1.4pt, error bars/.cd, y dir=both, y explicit] coordinates {
      (1000,69.006295) +- (0,1.076463)
      (2000,72.480278) +- (0,0.899097)
      (3000,71.357023) +- (0,1.398893)
      (4000,73.391459) +- (0,0.226578)
      (5000,74.196779) +- (0,1.523936)
      (6000,72.586719) +- (0,3.531555)
      (7000,74.497840) +- (0,1.765155)
      (8000,73.292104) +- (0,2.834853)
      (9000,70.491651) +- (0,3.058154)
      (10000,75.008118) +- (0,1.740362)
      (11000,73.312707) +- (0,0.309686)
      (12000,74.433806) +- (0,0.890145)
      (13000,73.796260) +- (0,1.126800)
      (14000,74.248417) +- (0,1.045936)
      (15000,72.858946) +- (0,1.964005)
      (16000,75.769168) +- (0,1.372435)
      (17000,76.036869) +- (0,1.523045)
      (18000,73.924714) +- (0,1.591590)
      (19000,76.344156) +- (0,0.235483)
      (20000,75.081931) +- (0,1.272039)
      (21000,74.521325) +- (0,0.432310)
      (22000,75.174929) +- (0,0.840219)
      (23000,75.771846) +- (0,0.568744)
      (24000,75.912737) +- (0,0.790963)
      (25000,74.032865) +- (0,1.364103)
      (26000,74.477717) +- (0,0.872947)
      (26050,76.217167) +- (0,1.220422)
    };
    \addplot[mark=triangle*, thick, gray!55!black, mark size=1.4pt, error bars/.cd, y dir=both, y explicit] coordinates {
      (1000,62.574370) +- (0,1.829127)
      (2000,58.545045) +- (0,10.256800)
      (3000,61.044647) +- (0,11.571197)
      (4000,56.995604) +- (0,16.576651)
      (5000,69.462891) +- (0,1.154414)
      (6000,71.945081) +- (0,1.212319)
      (7000,70.113968) +- (0,1.881118)
      (8000,71.149323) +- (0,1.174926)
      (9000,71.630148) +- (0,0.425274)
      (10000,69.992157) +- (0,2.786322)
      (11000,71.652077) +- (0,0.855550)
      (12000,68.660606) +- (0,2.099024)
      (13000,72.984384) +- (0,2.717095)
      (14000,71.626686) +- (0,0.493618)
      (15000,73.054341) +- (0,1.258797)
      (16000,70.357963) +- (0,1.163119)
      (17000,71.888701) +- (0,0.741512)
      (18000,72.437618) +- (0,1.281280)
      (19000,71.598466) +- (0,1.552200)
      (20000,72.769638) +- (0,0.638810)
      (21000,72.867202) +- (0,2.388309)
      (22000,71.662831) +- (0,0.824005)
      (23000,73.014350) +- (0,0.904818)
      (24000,71.711958) +- (0,0.657162)
      (25000,73.759607) +- (0,2.440518)
      (26000,72.018213) +- (0,1.414673)
      (26050,70.602347) +- (0,0.970579)
    };
    \addplot[dashed, thick, black, no markers] coordinates {(0,72.195818) (27000,72.195818)};

    \nextgroupplot[title={billiger.de}, xmin=0, xmax=6000, ymin=15, ymax=70, xtick={0,2000,4000,6000}]
    \addplot[mark=*, thick, blue!70!black, mark size=1.4pt, error bars/.cd, y dir=both, y explicit] coordinates {
      (1000,23.426158) +- (0,4.867468)
      (2000,54.223759) +- (0,3.301157)
      (3000,58.831715) +- (0,2.100122)
      (4000,60.650873) +- (0,1.136453)
      (5000,59.921366) +- (0,1.528278)
      (6000,59.997957) +- (0,1.311876)
    };
    \addplot[mark=square*, thick, orange!85!black, mark size=1.4pt, error bars/.cd, y dir=both, y explicit] coordinates {
      (1000,23.426158) +- (0,4.867468)
      (2000,45.394906) +- (0,4.694792)
      (3000,48.236023) +- (0,2.556063)
      (4000,50.181069) +- (0,1.196636)
      (5000,57.364631) +- (0,1.684184)
      (6000,56.283478) +- (0,1.361575)
    };
    \addplot[mark=triangle*, thick, gray!55!black, mark size=1.4pt, error bars/.cd, y dir=both, y explicit] coordinates {
      (1000,44.113801) +- (0,1.868880)
      (2000,51.025389) +- (0,3.027977)
      (3000,56.245337) +- (0,0.874890)
      (4000,56.472687) +- (0,1.512435)
      (5000,58.867626) +- (0,2.555694)
      (6000,58.845279) +- (0,2.698844)
    };
    \addplot[dashed, thick, black, no markers] coordinates {(0,65.086522) (6000,65.086522)};

    \end{groupplot}
    \path (budgetplots c1r1.north west) -- node[midway, left=25pt, rotate=90, anchor=center, font=\small] {Student model F1} (budgetplots c1r2.south west);
    \path (budgetplots c1r2.south west) -- node[midway, below=15pt, font=\small] {Training pairs (thousands)} (budgetplots c4r2.south east);
    \path (budgetplots c1r2.south west) -- node[midway, below=30pt] {\pgfplotslegendfromname{budgetlegend}} (budgetplots c4r2.south east);
  \end{tikzpicture}
  \caption{Ditto F1 by training-set size. Error bars: $\pm$ one SD.}
  \Description{Eight panels show student F1 versus training-pair budget for three selection strategies, with seed standard deviations and benchmark-trained baselines. The added panels are Semi-HETER, Dn7 Walmart-Amazon and billiger.de with 50 percent unseen entities.}
  \label{fig:al-budget}
\end{figure*}

On all benchmarks except billiger.de, the best machine-labeled set exceeds benchmark training or trails it by less than 2 F1 (Table~\ref{tab:selection-strategies}). It exceeds benchmark training on five benchmarks, by 0.68 to 10.15 F1, and trails it by 0.44 F1 on DBLP-ACM and 1.79 F1 on DBLP-Scholar. These seven include the nearly saturated DBLP tasks, the imbalanced Dn7, and WDC Products, whose test entities are absent from all training sets. Student F1 differs significantly from benchmark training on DBLP-Scholar, where all three sets trail by 1.79 to 2.10 F1 and runs vary little (SD at most 0.38), and on Semi-HETER. On Dn7, the benchmark-trained runs vary strongly (SD 3.93), so gains of up to 4.02 F1 are not significant.

\begin{table}[!t]
  \centering
  \caption{Ditto F1 by teacher on AL-Ditto sets. $\uparrow$/$\downarrow$: significantly above/below benchmark training.}
  \label{tab:llm-comparison}
  \fontsize{7.5pt}{9pt}\selectfont
  \setlength{\tabcolsep}{1pt}
  \begin{tabular*}{\columnwidth}{@{\extracolsep{\fill}}lrrrrr@{}}
    \toprule
    Benchmark & \shortstack{Benchmark\\training set} & GPT-5.2 & \shortstack{Qwen 3.6\\Plus} & \shortstack{Kimi\\K2.6} & $\Delta$ \\
    \midrule
    \multicolumn{6}{l}{\textit{Products}} \\
    \textsc{Abt-Buy} & 88.34\,$\pm$ 1.54 & \underline{89.02}\,\underline{$\pm$ 0.11} & 87.04\,$\pm$ 0.57 & \textbf{89.77}\,\textbf{$\pm$ 0.34} & $+1.43$ \\
    \begin{tabular}[c]{@{}l@{}}\textsc{Walmart-}\\\textsc{Amazon}\end{tabular} & 85.66\,$\pm$ 1.04 & \underline{87.25}\,\underline{$\pm$ 0.65} & 85.34\,$\pm$ 1.09 & \textbf{87.59}\,\textbf{$\pm$ 0.99} & $+1.93$ \\
    \begin{tabular}[c]{@{}l@{}}Dn7\\Walmart-\\Amazon\end{tabular} & 72.20\,$\pm$ 3.93 & \textbf{75.64}\,\textbf{$\pm$ 2.86} & 73.29\,$\pm$ 0.49 & \underline{74.11}\,\underline{$\pm$ 1.44} & $+3.44$ \\
    \textsc{WDC Products} & 71.23\,$\pm$ 0.83 & \textbf{72.17}\,\textbf{$\pm$ 1.01} & 71.88\,$\pm$ 1.31 & \underline{72.09}\,\underline{$\pm$ 0.73} & $+0.94$ \\
    billiger.de & \textbf{65.09}\,\textbf{$\pm$ 2.11} & 60.25\,$\pm$ 3.87 & \underline{62.67}\,\underline{$\pm$ 0.71} & 60.05\,$\pm$ 0.91 & $-2.42$ \\
    \midrule
    \multicolumn{6}{l}{\textit{Literature}} \\
    \textsc{DBLP-ACM} & \textbf{98.43}\,\textbf{$\pm$ 0.34} & \underline{97.91}\,\underline{$\pm$ 0.60} & 97.37\,$\pm$ 0.04 & 97.80\,$\pm$ 0.35 & $-0.52$ \\
    \begin{tabular}[c]{@{}l@{}}\textsc{DBLP-}\\\textsc{Scholar}\end{tabular} & \textbf{95.64}\,\textbf{$\pm$ 0.26} & \underline{93.86}$^{\downarrow}$\,\underline{$\pm$ 0.23} & 93.31$^{\downarrow}$\,$\pm$ 0.77 & 93.04$^{\downarrow}$\,$\pm$ 0.55 & $-1.78$ \\
    Semi-HETER & 82.97\,$\pm$ 5.29 & 91.53$^{\uparrow}$\,$\pm$ 2.24 & \underline{93.41}$^{\uparrow}$\,\underline{$\pm$ 0.84} & \textbf{93.61}$^{\uparrow}$\,\textbf{$\pm$ 1.90} & $+10.64$ \\
    \bottomrule
  \end{tabular*}
\end{table}

\begin{table*}[!b]
  \centering
  \caption{Ditto F1 after post-processing AL-Ditto training sets. $\uparrow$/$\downarrow$: significantly above/below the unprocessed AL-Ditto set.}
  \label{tab:ditto-refinements}
  \setlength{\tabcolsep}{2.5pt}
  \footnotesize
  \resizebox{\textwidth}{!}{
  \begin{tabular}{lrrrrrrrr}
    \toprule
    Benchmark & \shortstack{Benchmark\\training set} & AL-Ditto & +Relabel & +Relabel drop & +Closure drop & +Cl.\,$\wedge$\,Rel.\ drop & +Cl.\,$\vee$\,Rel.\ drop & $\Delta$ to bench. \\
    \midrule
    \multicolumn{9}{l}{\textit{Products}} \\
    \textsc{Abt-Buy} & 88.34 $\pm$ 1.54 & 89.02 $\pm$ 0.11 & \textbf{90.84 $\pm$ 0.46} & 89.73 $\pm$ 0.84 & 89.38 $\pm$ 0.70 & 89.89 $\pm$ 1.19 & \underline{90.21$^{\uparrow}$ $\pm$ 0.21} & $+2.50$ \\
    \textsc{Walmart-Amazon} & 85.66 $\pm$ 1.04 & 87.25 $\pm$ 0.65 & 84.42 $\pm$ 1.53 & \textbf{87.97 $\pm$ 0.75} & 87.47 $\pm$ 0.52 & 87.69 $\pm$ 0.63 & \underline{87.85 $\pm$ 0.64} & $+2.31$ \\
    Dn7 Walmart-Amazon & 72.20 $\pm$ 3.93 & \textbf{75.64 $\pm$ 2.86} & 73.05 $\pm$ 1.39 & 74.82 $\pm$ 0.34 & 75.49 $\pm$ 1.56 & \underline{75.59 $\pm$ 0.71} & 75.55 $\pm$ 0.74 & $+3.44$ \\
    \textsc{WDC Products} & 71.23 $\pm$ 0.83 & \underline{72.17 $\pm$ 1.01} & \textbf{73.35 $\pm$ 1.30} & 71.40 $\pm$ 1.07 & 69.80 $\pm$ 1.97 & 71.75 $\pm$ 1.06 & 69.48 $\pm$ 1.05 & $+2.12$ \\
    billiger.de & \textbf{65.09 $\pm$ 2.11} & 60.25 $\pm$ 3.87 & 59.81 $\pm$ 2.28 & \underline{63.71 $\pm$ 1.16} & 55.89 $\pm$ 2.11 & 63.00 $\pm$ 1.72 & 61.17 $\pm$ 1.63 & $-1.38$ \\
    \midrule
    \multicolumn{9}{l}{\textit{Literature}} \\
    \textsc{DBLP-ACM} & \textbf{98.43 $\pm$ 0.34} & \underline{97.91 $\pm$ 0.60} & 96.91 $\pm$ 0.24 & 97.62 $\pm$ 0.32 & \underline{97.91 $\pm$ 0.13} & 97.69 $\pm$ 0.06 & 97.83 $\pm$ 0.48 & $-0.52$ \\
    \textsc{DBLP-Scholar} & \textbf{95.64 $\pm$ 0.26} & \underline{93.86 $\pm$ 0.23} & 93.78 $\pm$ 0.68 & 93.71 $\pm$ 0.26 & 92.79$^{\downarrow}$ $\pm$ 0.08 & 93.84 $\pm$ 0.21 & 93.60 $\pm$ 0.21 & $-1.78$ \\
    Semi-HETER & 82.97 $\pm$ 5.29 & 91.53 $\pm$ 2.24 & 91.20 $\pm$ 1.06 & \underline{92.87 $\pm$ 2.61} & 90.46 $\pm$ 2.16 & 87.52 $\pm$ 6.97 & \textbf{94.76 $\pm$ 1.50} & $+11.78$ \\
    \bottomrule
  \end{tabular}}
\end{table*}

AL-Ditto has the highest mean F1 of the three strategies on five benchmarks. AL-ML leads on Dn7 and Semi-HETER and similarity search on DBLP-ACM, with AL-Ditto at most 1.60 F1 behind in these three cases. The strategies differ by less than 0.50 F1 on the two DBLP tasks and by 5.61 F1 on Dn7, where both active-learning sets exceed similarity search (Table~\ref{tab:selection-strategies}). Without choosing the best strategy per task, AL-Ditto alone stays at most 1.78 F1 below benchmark training or above it on seven of eight tasks.

On Semi-HETER, all three sets, including similarity search, yield students significantly above benchmark training, by 8.35 to 10.15 F1. Section~\ref{sec:composition} relates this gain to set composition.

On billiger.de, Ditto trained on any of the three sets remains 5.09 to 8.80 F1 below benchmark training, with a significant difference for AL-ML (Table~\ref{tab:selection-strategies}). The benchmark-trained and machine-labeled students share the same RoBERTa encoder, so the student's language coverage alone cannot explain this relative gap. Lower teacher label accuracy on these German offers is one contributor. GPT-5.2 labels only 80.71\% of the benchmark's matching training pairs as matches, the lowest recall of the eight benchmarks, and the following control quantifies the effect.

\textbf{Labels versus pairs.} The machine-labeled sets change both the pairs and the labels. To separate the two, we keep the benchmark training pairs and replace only their labels with GPT-5.2 labels. Replacing the labels changes mean F1 by $-2.45$ to $+0.77$, and the change is significant only on DBLP-Scholar. There, the relabeled pairs lose 2.38 F1 and the three selected sets 1.79 to 2.10 F1 (Table~\ref{tab:selection-strategies}), so the gap on this task is a label effect. On the other seven benchmarks, the best selected set scores above the relabeled benchmark pairs, from 0.16 F1 on DBLP-ACM and 0.60 F1 on DBLP-Scholar to 3.49 F1 on Dn7 and 9.39 F1 on Semi-HETER, where run-to-run variation is large. Pair selection can therefore contributes to fitness for use beyond the labels. On billiger.de, the relabeled pairs lose 2.45 F1 and the selected sets a further 2.64 to 6.36 F1, so teacher labels and selection both contribute to the gap on this task.

\noindent\textbf{Finding.} Machine-labeled sets from GPT-5.2 are a practical alternative to benchmark training for the PLM-based Ditto on seven of eight tasks, including hard non-matches, unseen entities, and structural variations. On the German offers of billiger.de, teacher labels and pair selection both leave a gap (Table~\ref{tab:selection-strategies}).

\subsection{Effect of the Label Budget}
\label{sec:budget}

We examine how student F1 scales with the number of labeled pairs by applying the three pair-selection strategies under increasing label budgets, with GPT-5.2 as teacher (Figure~\ref{fig:al-budget}). The benchmark reference is a single full-size training set, so the curves compare the strategies with each other and with this fixed reference.

At small budgets, AL-Ditto often reaches a given F1 with fewer labels (Figure~\ref{fig:al-budget}). On Abt-Buy, it reaches 88.06 F1 at 2,000 pairs, which neither AL-ML nor similarity search reaches within 6,000 pairs. On billiger.de, AL-Ditto leads from 2,000 to 6,000 pairs and reaches 58.83 F1 at 3,000 pairs, which similarity search reaches only at 5,000 pairs and AL-ML not at all. On Dn7, AL-Ditto reaches 74.42 F1 at 2,000 pairs, above similarity search at every budget and above the benchmark-trained student with 26,050 pairs. Its lead is limited to the first 2,000 pairs on Walmart-Amazon, and on WDC Products it leads at most budgets from 2,000 pairs on, by up to 3.35 F1. Similarity search leads at the smallest budget on WDC Products, billiger.de, and Semi-HETER, where it reaches 92.73 F1 at 600 pairs.

Larger budgets narrow the differences between strategies only partly. At the largest budget, the three strategies lie within 1.80 F1 of each other on Walmart-Amazon, Semi-HETER, and the two DBLP tasks, down from up to 11.74 F1 at small budgets. Gaps persist on the other four benchmarks. AL-Ditto leads by up to 2.25 F1 on Abt-Buy at 6,000 pairs and 2.64 F1 on WDC Products at 19,000 pairs, and similarity search ends 5.04 F1 below AL-Ditto on Dn7. On billiger.de, similarity search closes to within 1.15 F1 of AL-Ditto at 6,000 pairs, while AL-ML stays 3.71 F1 behind.

The budget at which a curve flattens depends on the task. All three strategies are within 2 F1 of their final DBLP level at 1,000 pairs. WDC Products gains until about 15,000 pairs, and AL-Ditto is the only strategy to reach the benchmark reference, near 13,000 pairs. On billiger.de, all three curves stay below the benchmark reference, and AL-Ditto remains near 60 F1 from 4,000 pairs on, so additional labels alone do not close this gap.

\noindent\textbf{Finding.} Pair selection matters most at small label budgets, where AL-Ditto often needs fewer labels, and on the hard non-matches of Dn7, where similarity search trails AL-Ditto even at the largest budget. Larger budgets narrow the gaps on several tasks (Figure~\ref{fig:al-budget}).

\subsection{Effect of the Teacher Model}
\label{sec:labeler}
This section examines how the choice of teacher affects fitness for use. We rerun the AL-Ditto workflow with two alternative teachers, the proprietary Qwen 3.6 Plus and the open-weight Kimi K2.6, and keep the selection strategy, student training, and evaluation fixed (Table~\ref{tab:llm-comparison}). An open-weight teacher can run locally, which matters for applications that cannot share their records with external model providers. As teacher API charges are low (Section~\ref{sec:cost}), we do not evaluate smaller and cheaper teachers.

All three teachers show the significance pattern of Section~\ref{sec:f1-comparison}. Student F1 differs significantly from benchmark training only on Semi-HETER, where all three teachers exceed it by 8.56 to 10.64 F1, and on DBLP-Scholar, where they remain 1.78 to 2.60 F1 below (Table~\ref{tab:llm-comparison}). On the other six, no teacher differs significantly.

Kimi K2.6 has the highest mean student F1 on three benchmarks, GPT-5.2 on four, and Qwen on billiger.de. The spread among the teachers is below 0.9 F1 on WDC Products and the two DBLP tasks and 2.08 to 2.73 F1 on the other five. Kimi and GPT-5.2 do not differ significantly on any benchmark, and Kimi's mean trails GPT-5.2's by at most 1.53 F1, on Dn7. The only significant differences between two teachers are on Abt-Buy, where Kimi leads Qwen by 2.73 F1 and GPT-5.2 leads Qwen by 1.98 F1. For billiger.de, the GPT-5.2 column uses the 5,897-pair AL-Ditto set of Section~\ref{sec:post-processing}.

\noindent\textbf{Finding.} The open-weight Kimi K2.6 trains Ditto students that are competitive with those trained on GPT-5.2 labels, so the workflow does not depend on a proprietary teacher (Table~\ref{tab:llm-comparison}).

\subsection{Cleaning the Machine-labeled Training Sets}
\label{sec:post-processing}
\begingroup\setlength{\emergencystretch}{1em}
This section evaluates how post-pro\-cessing the teacher labels changes Ditto F1. We apply the relabeling and closure-based methods from Section~\ref{sec:method} to the AL-Ditto training sets, separately and in combination. On billiger.de, post-processing starts from the 5,897-pair AL-Ditto set, so its reference differs slightly from Table~\ref{tab:selection-strategies}. We test each variant against the unprocessed AL-Ditto set as in Section~\ref{sec:setup}, correcting over the five variants per benchmark. Relabeling preserves the selected pairs, while dropping changes the amount and composition of the training data. In Table~\ref{tab:ditto-refinements}, $\wedge$ drops pairs flagged by both checks and $\vee$ pairs flagged by either check.\par\endgroup

Two of 40 post-processed cells differ significantly from the unprocessed set: the union filter raises Abt-Buy by 1.19 F1 and closure alone lowers DBLP-Scholar by 1.07 (Table~\ref{tab:ditto-refinements}). None raises mean F1 everywhere. Relabeling raises F1 by 1.82 on Abt-Buy and 1.18 on WDC Products but lowers it by 2.83 on Walmart-Amazon, where relabel-drop gains 0.72. On the DBLP tasks, no variant helps.

Closure filtering alone raises mean F1 on Abt-Buy (+0.36) and Walmart-Amazon (+0.22), leaves DBLP-ACM unchanged, and lowers it on the other five, most on billiger.de ($-4.36$), where it also removes 87 conflicting negative pairs. Combined with relabeling, the intersection raises billiger.de by 2.75, and the union raises Abt-Buy (+1.19), Walmart-Amazon (+0.60), billiger.de (+0.92), and Semi-HETER (+3.23), where it reaches the highest F1 of all variants. Sparse positive graphs may contain correct bridges and negatives, so the filter can remove useful supervision along with erroneous links.

\noindent\textbf{Finding.} Each post-processing method raises student F1 on some tasks, but none on all, and most changes are not significant.

\subsection{Performance of Different Student Models}
\label{sec:student-models}

All previous experiments use Ditto as the student. To examine how the utility of a training source depends on the student, we repeat the comparison with the XGBoost and Qwen3 students (Section~\ref{sec:setup}). Table~\ref{tab:student-models} reports each student on every training source, with zero-shot results for Qwen3, the only family that matches without training.

\begin{table*}[!t]
  \centering
  \caption{F1 by student and training source. $\uparrow$/$\downarrow$: significantly above/below benchmark training of the same student.}
  \label{tab:student-models}
  \small
  \setlength{\tabcolsep}{2pt}
  \begin{tabular*}{0.85\textwidth}{@{\extracolsep{\fill}}llccccc@{}}
    \toprule
    Benchmark & Model & Zero-shot & \shortstack{Benchmark\\training set} & Sim. search & AL-ML & AL-Ditto \\
    \midrule
    \multicolumn{7}{l}{\textit{Products}} \\
    \multirow{5}{*}{\textsc{Abt-Buy}}        & XGBoost    & n/a & 45.54 $\pm$ 0.53 & 52.97$^{\uparrow}$ $\pm$ 0.58 & 64.83$^{\uparrow}$ $\pm$ 0.60 & 55.49$^{\uparrow}$ $\pm$ 0.44 \\
                                             & Ditto      & n/a & 88.34 $\pm$ 1.54 & 86.77 $\pm$ 0.09 & 87.76 $\pm$ 0.67 & 89.02 $\pm$ 0.11 \\
                                             & Qwen3-0.6B & 35.74 & 86.17 $\pm$ 0.13 & 84.29$^{\downarrow}$ $\pm$ 0.24 & 86.63 $\pm$ 0.65 & 88.42$^{\uparrow}$ $\pm$ 0.60 \\
                                             & Qwen3-1.7B & 50.61 & 88.32 $\pm$ 1.68 & 85.04 $\pm$ 0.83 & 87.42 $\pm$ 0.06 & 88.44 $\pm$ 0.35 \\
                                             & Qwen3-8B   & 85.31 & \textbf{92.05 $\pm$ 0.31} & 86.65$^{\downarrow}$ $\pm$ 1.05 & 89.45$^{\downarrow}$ $\pm$ 0.15 & \underline{89.58$^{\downarrow}$ $\pm$ 1.08} \\
    \midrule
    \multirow{5}{*}{\textsc{Walmart-Amazon}} & XGBoost    & n/a & 61.13 $\pm$ 3.08 & 60.07 $\pm$ 0.77 & 61.11 $\pm$ 1.26 & 57.09 $\pm$ 1.13 \\
                                             & Ditto      & n/a & 85.66 $\pm$ 1.04 & 85.37 $\pm$ 1.55 & 86.88 $\pm$ 0.43 & 87.25 $\pm$ 0.65 \\
                                             & Qwen3-0.6B & 24.45 & 84.21 $\pm$ 0.62 & 85.18 $\pm$ 1.42 & 84.33 $\pm$ 0.65 & 84.56 $\pm$ 0.71 \\
                                             & Qwen3-1.7B & 40.30 & 85.84 $\pm$ 1.91 & 84.26 $\pm$ 2.39 & 86.27 $\pm$ 0.74 & 85.24 $\pm$ 0.33 \\
                                             & Qwen3-8B   & 73.33 & \textbf{90.33 $\pm$ 0.13} & 87.05$^{\downarrow}$ $\pm$ 0.20 & \underline{88.27$^{\downarrow}$ $\pm$ 0.50} & 87.91$^{\downarrow}$ $\pm$ 0.31 \\
    \midrule
    \multirow{5}{*}{Dn7 Walmart-Amazon} & XGBoost & n/a & 58.82 $\pm$ 0.82 & 59.02 $\pm$ 0.31 & 64.51$^{\uparrow}$ $\pm$ 0.09 & 62.70$^{\uparrow}$ $\pm$ 0.20 \\
     & Ditto & n/a & 72.20 $\pm$ 3.93 & 70.60 $\pm$ 0.97 & \textbf{76.22 $\pm$ 1.22} & 75.64 $\pm$ 2.86 \\
     & Qwen3-0.6B & 12.97 & 55.42 $\pm$ 5.90 & 65.23 $\pm$ 7.69 & 68.49 $\pm$ 5.41 & 71.40 $\pm$ 2.32 \\
     & Qwen3-1.7B & 19.23 & 64.41 $\pm$ 0.18 & 66.54 $\pm$ 3.12 & 72.12 $\pm$ 2.83 & 71.94 $\pm$ 2.47 \\
     & Qwen3-8B & 57.98 & 75.61 $\pm$ 2.93 & 70.30 $\pm$ 1.80 & \underline{76.08 $\pm$ 1.00} & 74.18 $\pm$ 1.84 \\
    \midrule
    \multirow{5}{*}{\textsc{WDC Products}}            & XGBoost    & n/a & 36.61 $\pm$ 0.73 & 37.24 $\pm$ 0.24 & 36.40 $\pm$ 0.61 & 36.95 $\pm$ 0.05 \\
                                             & Ditto      & n/a & 71.23 $\pm$ 0.83 & 69.53 $\pm$ 1.25 & 70.65 $\pm$ 0.42 & 72.17 $\pm$ 1.01 \\
                                             & Qwen3-0.6B & 40.95 & 57.14 $\pm$ 1.17 & 53.94 $\pm$ 1.72 & 60.19 $\pm$ 0.61 & 58.14 $\pm$ 1.78 \\
                                             & Qwen3-1.7B & 22.59 & 67.05 $\pm$ 1.27 & 61.92 $\pm$ 4.69 & 65.10 $\pm$ 0.32 & 56.77 $\pm$ 4.63 \\
                                             & Qwen3-8B   & 66.14 & \textbf{73.53 $\pm$ 2.84} & 71.19 $\pm$ 6.89 & 67.91 $\pm$ 1.59 & \underline{72.92 $\pm$ 1.63} \\
    \midrule
    \multirow{5}{*}{billiger.de} & XGBoost & n/a & 34.07 $\pm$ 0.10 & 32.94$^{\downarrow}$ $\pm$ 0.30 & 32.33 $\pm$ 0.67 & 34.27 $\pm$ 0.13 \\
     & Ditto & n/a & \textbf{65.09 $\pm$ 2.11} & 58.85 $\pm$ 2.70 & 56.28$^{\downarrow}$ $\pm$ 1.36 & 60.00 $\pm$ 1.31 \\
     & Qwen3-0.6B & 29.79 & 38.88 $\pm$ 0.91 & 45.62 $\pm$ 6.50 & 48.83 $\pm$ 4.35 & 42.79 $\pm$ 1.37 \\
     & Qwen3-1.7B & 32.17 & 33.74 $\pm$ 6.01 & 51.94 $\pm$ 0.89 & 52.91 $\pm$ 0.17 & 51.02 $\pm$ 1.52 \\
     & Qwen3-8B & 57.08 & 43.91 $\pm$ 1.33 & \underline{64.47$^{\uparrow}$ $\pm$ 0.61} & 61.45$^{\uparrow}$ $\pm$ 0.93 & 55.67$^{\uparrow}$ $\pm$ 2.41 \\
    \midrule
    \multicolumn{7}{l}{\textit{Literature}} \\
    \multirow{5}{*}{\textsc{DBLP-ACM}}       & XGBoost    & n/a & 92.76 $\pm$ 3.33 & 96.85 $\pm$ 0.16 & 96.72 $\pm$ 0.07 & 96.70 $\pm$ 0.00 \\
                                             & Ditto      & n/a & 98.43 $\pm$ 0.34 & 97.99 $\pm$ 0.40 & 97.50 $\pm$ 0.17 & 97.91 $\pm$ 0.60 \\
                                             & Qwen3-0.6B & 78.65 & 98.17 $\pm$ 0.33 & 97.17$^{\downarrow}$ $\pm$ 0.16 & 97.10$^{\downarrow}$ $\pm$ 0.13 & 96.99$^{\downarrow}$ $\pm$ 0.13 \\
                                             & Qwen3-1.7B & 79.48 & \textbf{98.61 $\pm$ 0.13} & 97.25$^{\downarrow}$ $\pm$ 0.21 & 97.39$^{\downarrow}$ $\pm$ 0.06 & 96.88$^{\downarrow}$ $\pm$ 0.06 \\
                                             & Qwen3-8B   & 91.54 & \underline{98.46 $\pm$ 1.29} & 98.01 $\pm$ 0.11 & 97.97 $\pm$ 0.06 & 98.00 $\pm$ 0.11 \\
    \midrule
    \multirow{5}{*}{\textsc{DBLP-Scholar}}   & XGBoost    & n/a & 55.01 $\pm$ 2.60 & 76.33$^{\uparrow}$ $\pm$ 1.47 & 70.39$^{\uparrow}$ $\pm$ 4.34 & 76.67$^{\uparrow}$ $\pm$ 0.80 \\
                                             & Ditto      & n/a & \textbf{95.64 $\pm$ 0.26} & 93.54$^{\downarrow}$ $\pm$ 0.18 & 93.81$^{\downarrow}$ $\pm$ 0.38 & 93.86$^{\downarrow}$ $\pm$ 0.23 \\
                                             & Qwen3-0.6B & 67.89 & 94.72 $\pm$ 0.27 & 92.29$^{\downarrow}$ $\pm$ 0.27 & 92.43$^{\downarrow}$ $\pm$ 0.62 & 91.30 $\pm$ 1.75 \\
                                             & Qwen3-1.7B & 56.40 & 95.10 $\pm$ 0.04 & 91.70$^{\downarrow}$ $\pm$ 0.38 & 92.30$^{\downarrow}$ $\pm$ 0.60 & 92.33$^{\downarrow}$ $\pm$ 0.07 \\
                                             & Qwen3-8B   & 84.34 & \underline{95.12 $\pm$ 0.12} & 92.32$^{\downarrow}$ $\pm$ 0.16 & 91.97$^{\downarrow}$ $\pm$ 0.25 & 92.52$^{\downarrow}$ $\pm$ 0.22 \\
    \midrule
    \multirow{5}{*}{Semi-HETER} & XGBoost & n/a & 82.55 $\pm$ 0.47 & 89.29$^{\uparrow}$ $\pm$ 0.79 & 81.03 $\pm$ 0.67 & 81.64 $\pm$ 1.45 \\
     & Ditto & n/a & 82.97 $\pm$ 5.29 & 91.33$^{\uparrow}$ $\pm$ 1.92 & 93.13$^{\uparrow}$ $\pm$ 3.62 & 91.53$^{\uparrow}$ $\pm$ 2.24 \\
     & Qwen3-0.6B & 78.09 & 86.48 $\pm$ 1.01 & \textbf{94.21$^{\uparrow}$ $\pm$ 0.25} & 89.62 $\pm$ 2.66 & 93.01$^{\uparrow}$ $\pm$ 0.36 \\
     & Qwen3-1.7B & 62.07 & 81.35 $\pm$ 4.73 & 90.89 $\pm$ 1.80 & 50.23$^{\downarrow}$ $\pm$ 9.02 & 61.89$^{\downarrow}$ $\pm$ 1.70 \\
     & Qwen3-8B & 83.64 & 93.02 $\pm$ 0.62 & 92.55 $\pm$ 1.67 & 81.90$^{\downarrow}$ $\pm$ 1.58 & \underline{93.67 $\pm$ 0.42} \\
    \bottomrule
  \end{tabular*}
\end{table*}

\begin{table*}[!t]
  \centering
  \caption{Test F1 of each zero-shot teacher (one run) and of Ditto trained on its AL-Ditto labels (mean $\pm$ SD, from Table~\ref{tab:llm-comparison}). $\Delta$: teacher minus own-student F1, positive when the student trails.}
  \label{tab:llm-direct}
  \footnotesize
  \setlength{\tabcolsep}{2pt}
  \begin{tabular*}{\textwidth}{@{\extracolsep{\fill}}lrrrrrrrrrr@{}}
    \toprule
     & & \multicolumn{3}{c}{GPT-5.2} & \multicolumn{3}{c}{Qwen 3.6 Plus} & \multicolumn{3}{c}{Kimi K2.6} \\
    \cmidrule(lr){3-5}\cmidrule(lr){6-8}\cmidrule(lr){9-11}
    Benchmark & \shortstack{Benchmark-\\trained Ditto} & Direct & Ditto & $\Delta$ & Direct & Ditto & $\Delta$ & Direct & Ditto & $\Delta$ \\
    \midrule
    \multicolumn{11}{l}{\textit{Products}} \\
    \textsc{Abt-Buy} & 88.34\,{\scriptsize $\pm$\,1.54} & 91.53 & 89.02\,{\scriptsize $\pm$\,0.11} & $+2.51$ & 92.10 & 87.04\,{\scriptsize $\pm$\,0.57} & $+5.06$ & 93.81 & 89.77\,{\scriptsize $\pm$\,0.34} & $+4.04$ \\
    \textsc{Walmart-Amazon} & 85.66\,{\scriptsize $\pm$\,1.04} & 88.78 & 87.25\,{\scriptsize $\pm$\,0.65} & $+1.53$ & 85.71 & 85.34\,{\scriptsize $\pm$\,1.09} & $+0.37$ & 90.54 & 87.59\,{\scriptsize $\pm$\,0.99} & $+2.95$ \\
    Dn7 Walmart-Amazon & 72.20\,{\scriptsize $\pm$\,3.93} & 76.84 & 75.64\,{\scriptsize $\pm$\,2.86} & $+1.20$ & 75.63 & 73.29\,{\scriptsize $\pm$\,0.49} & $+2.34$ & 80.00 & 74.11\,{\scriptsize $\pm$\,1.44} & $+5.89$ \\
    \textsc{WDC Products} & 71.23\,{\scriptsize $\pm$\,0.83} & 84.47 & 72.17\,{\scriptsize $\pm$\,1.01} & $+12.30$ & 86.91 & 71.88\,{\scriptsize $\pm$\,1.31} & $+15.03$ & 87.20 & 72.09\,{\scriptsize $\pm$\,0.73} & $+15.11$ \\
    billiger.de & 65.09\,{\scriptsize $\pm$\,2.11} & 75.71 & 60.25\,{\scriptsize $\pm$\,3.87} & $+15.46$ & 69.11 & 62.67\,{\scriptsize $\pm$\,0.71} & $+6.44$ & 76.12 & 60.05\,{\scriptsize $\pm$\,0.91} & $+16.07$ \\
    \midrule
    \multicolumn{11}{l}{\textit{Literature}} \\
    \textsc{DBLP-ACM} & 98.43\,{\scriptsize $\pm$\,0.34} & 96.82 & 97.91\,{\scriptsize $\pm$\,0.60} & $-1.09$ & 93.87 & 97.37\,{\scriptsize $\pm$\,0.04} & $-3.50$ & 97.05 & 97.80\,{\scriptsize $\pm$\,0.35} & $-0.75$ \\
    \textsc{DBLP-Scholar} & 95.64\,{\scriptsize $\pm$\,0.26} & 90.87 & 93.86\,{\scriptsize $\pm$\,0.23} & $-2.99$ & 88.96 & 93.31\,{\scriptsize $\pm$\,0.77} & $-4.35$ & 92.49 & 93.04\,{\scriptsize $\pm$\,0.55} & $-0.55$ \\
    Semi-HETER & 82.97\,{\scriptsize $\pm$\,5.29} & 90.41 & 91.53\,{\scriptsize $\pm$\,2.24} & $-1.12$ & 95.24 & 93.41\,{\scriptsize $\pm$\,0.84} & $+1.83$ & 94.50 & 93.61\,{\scriptsize $\pm$\,1.90} & $+0.89$ \\
    \bottomrule
  \end{tabular*}
\end{table*}

Ditto is competitive with much larger students. With about 125 million parameters, it is more than 60 times smaller than Qwen3-8B. We compare the best trained result per model, the maximum over benchmark training and the three machine-labeled sources (Table~\ref{tab:student-models}). Ditto has the highest such result of all students on Dn7 (76.22 F1) and billiger.de (65.09 F1). It lies within 1.36 F1 of Qwen3-8B on six of the eight benchmarks and above it on three. Qwen3-8B leads by about 3 F1 only on Abt-Buy and Walmart-Amazon, at 47.12 times Ditto's inference time in our setup (Section~\ref{sec:cost}).

Whether a machine-labeled source matches benchmark training depends on the student. Significant losses concentrate on the two near-saturated DBLP tasks, where 14 of the 18 Qwen3 cells trail benchmark training significantly, by 1.00 to 3.40 F1, and on Abt-Buy and Walmart-Amazon for Qwen3-8B, whose benchmark-trained runs reach the highest F1 of all students there. Significant gains occur on Semi-HETER for Ditto and Qwen3-0.6B, for XGBoost on four benchmarks, and for Qwen3-8B on billiger.de. Source rankings also differ between students. Semi-HETER AL-ML yields 93.13 F1 with Ditto but 50.23 with Qwen3-1.7B, without response-parsing failures. These Qwen3-1.7B runs all select the first evaluation checkpoint and reach high precision at 26 to 44\% recall. The benchmark-trained Qwen3-8B runs on billiger.de stop after about 2.2 to 2.4 epochs and score 43.91 F1, below the zero-shot 57.08. Both low results coincide with early checkpoint selection, an association that does not establish whether optimization or the training data causes them.

XGBoost benefits from machine-labeled sets but remains weaker in absolute terms. Its best source is within 0.02 F1 of, or above, benchmark training on all eight tasks, while its mean F1 across benchmarks and sources is 61.88 versus 82.60 for Ditto. The best XGBoost result trails the best Ditto result by 1.58 F1 on DBLP-ACM and 3.84 F1 on Semi-HETER, but by 11.71 to 34.93 F1 on the other six benchmarks. Reproducing a weak benchmark-trained reference therefore does not establish competitive matching performance.

\noindent\textbf{Finding.} The machine-labeled sets are fit for use for all three student families on most tasks: the feature-based XGBoost, the PLM-based Ditto, and fine-tuned Qwen3 LLMs. Absolute F1 depends more on the student than on the training source, and the best source differs by student (Table~\ref{tab:student-models}).

\subsection{Performance of the Teacher Models}
\label{sec:direct-llm-matching}

We evaluate each teacher zero-shot on the test sets and compare it with its AL-Ditto student (Table~\ref{tab:llm-comparison}), with benchmark-trained Ditto as reference (Table~\ref{tab:llm-direct}). $\Delta$ is teacher F1 minus mean student F1, positive when the student trails. Each teacher result comes from a single zero-shot run, so the comparison is descriptive.

Each teacher's Ditto student exceeds its teacher on both DBLP tasks, by 0.55 to 4.35 F1 (Table~\ref{tab:llm-direct}). A possible explanation is that fine-tuning aggregates the teacher's decisions across many pairs into a consistent matching rule, which may smooth out isolated label errors. The GPT-5.2 student trails its teacher by 1.20 to 2.51 F1 on Abt-Buy, Walmart-Amazon, and Dn7 and leads it by 1.12 F1 on Semi-HETER, where the Qwen 3.6 Plus and Kimi K2.6 students trail their teachers by 1.83 and 0.89 F1.

The largest gaps occur on WDC Products and billiger.de, where each teacher leads the Ditto student trained on its own labels by 6.44 to 16.07 F1 (Table~\ref{tab:llm-direct}). Benchmark-trained Ditto also trails all teachers, by 13.24 to 15.97 F1 on WDC Products and 4.02 to 11.03 F1 on billiger.de. On WDC Products all test products and on billiger.de half of them are unseen during supervised training. PLM-based matchers lose accuracy on unseen products~\cite{peeters2024wdcproducts}, whereas LLMs are more robust to them~\cite{peeters2025llmem}, and our results on these two benchmarks reproduce this pattern. The training sets cover only about 500 products each (Section~\ref{sec:composition}).  Student F1 levels off as more pairs among these products are labeled, with AL-Ditto near 72 F1 on WDC Products from about 15,000 pairs and near 60 F1 on billiger.de from 4,000 pairs (Figure~\ref{fig:al-budget}).

\noindent\textbf{Finding.} The GPT-5.2 student is close to or above its teacher on most tasks. On WDC Products and billiger.de, with high shares of unseen products, the distilled and benchmark-trained PLM-based student fall behind, confirming the problems of PLM-based matchers with unseen entities reported by Peeters et~al.~\cite{peeters2025llmem} and Wang et~al.~\cite{wang2022bridging} (Table~\ref{tab:llm-direct}).

\subsection{Construction Cost and Inference Scaling}
\label{sec:cost}
\label{sec:llm-inference-scaling}
We relate the one-time cost of constructing a training set to the recurring runtime of the trained matcher. Table~\ref{tab:labeling-costs} reports all-query API costs and committee training time, Table~\ref{tab:inference-scaling} quantifies inference time, and Section~\ref{sec:direct-llm-matching} gives the accuracy trade-off.

\paragraph{Training-set construction.} Labeling all queried pairs of the eight benchmarks with GPT-5.2 cost \$64.85 for similarity search, \$46.92 for AL-ML, and \$49.37 for AL-Ditto at standard rates of \$1.75 per million input and \$14 per million output tokens\footnote{GPT-5.2 list prices without Batch or cache discounts at the time of the experiments: \url{https://developers.openai.com/api/docs/models/gpt-5.2}.} (Table~\ref{tab:labeling-costs}). These totals count every teacher call, including discarded pairs and seed labels. Manual labeling of the eight training sets would take 746.2 hours at 30 seconds per pair, or 248.7 hours at 10 seconds.

AL-Ditto adds local compute. Its committee fits sum to 157.30 single-GPU hours over all rounds, 81.49 of them on WDC Products and 16.92 on Dn7, counting concurrent fits separately.

\paragraph{Inference runtime.}
Table~\ref{tab:inference-scaling} averages normalized runtimes over eight benchmarks and projects linearly to one million pairs. Local measurements average three post-warmup passes on one A40 GPU with six CPU threads. Model loading is excluded, while neural tokenization, output processing, and XGBoost test-feature extraction are included. The API teachers run with ten concurrent workers, GPT-5.2 through the OpenAI API and Qwen 3.6 Plus and Kimi K2.6 through OpenRouter. Their averages combine complete current measurements on three benchmarks with historical measurements or estimates on the other five. Timing details are in the artifact.

Ditto averages 2.64 seconds per 1,000 pairs. Qwen3 students take 1.53 to 2.07 minutes (34.7 to 47.1 times longer), and API teachers take 1.49 to 20.17 minutes (33.9 to 458.8 times). GPT-5.2 with ten workers is slightly faster than the Qwen students evaluated one pair at a time, an ordering that depends on batching and concurrency.

\begin{table}[t]
\centering
  \caption{GPT-5.2 API cost (USD) and AL-Ditto committee training time.}
  \label{tab:labeling-costs}
  \small
  \setlength{\tabcolsep}{2.5pt}
  \begin{tabular*}{\columnwidth}{@{\extracolsep{\fill}}lrrrr@{}}
    \toprule
    Benchmark & \shortstack{Sim.\\search} & AL-ML & AL-Ditto & \shortstack{AL-Ditto\\fit h} \\
    \midrule
\multicolumn{5}{l}{\textit{Products}} \\
Abt-Buy & \$3.69 & \$2.81 & \$2.84 & 4.73 \\
Walmart-Amazon & \$5.53 & \$3.33 & \$2.69 & 3.10 \\
Dn7 Walmart-Amazon & \$12.45$^{*}$ & \$12.53$^{*}$ & \$12.42$^{*}$ & 16.92 \\
WDC Products & \$13.54 & \$10.83 & \$13.13 & 81.49 \\
billiger.de & \$6.41 & \$3.41$^{*}$ & \$6.77 & 30.23 \\
\midrule
\multicolumn{5}{l}{\textit{Literature}} \\
DBLP-ACM & \$6.66 & \$3.98 & \$3.00 & 2.89 \\
DBLP-Scholar & \$11.90 & \$9.22 & \$7.50 & 16.35 \\
Semi-HETER & \$4.67 & \$0.81 & \$1.02 & 1.60 \\
\midrule
Total & \$64.85$^{*}$ & \$46.92$^{*}$ & \$49.37$^{*}$ & 157.30 \\
    \bottomrule
  \end{tabular*}
  \par\smallskip
  \begin{minipage}{\columnwidth}\footnotesize
  Every teacher call is charged at standard rates. $^{*}$Partly estimated token usage. Fit h: summed committee training time, excluding final student fits.
  \end{minipage}
\end{table}

\begin{table}[t]
\centering
  \caption{Mean inference time across eight benchmarks and linear 1M-pair projections.}
  \label{tab:inference-scaling}
  \small
  \setlength{\tabcolsep}{2.5pt}
  \begin{tabular*}{\columnwidth}{@{\extracolsep{\fill}}lrrr@{}}
    \toprule
    Matcher & 1k pairs & \shortstack{1M pairs\\projection} & vs. Ditto \\
    \midrule
    \multicolumn{4}{l}{\textit{Trained students (Qwen: LoRA)}} \\
    XGBoost & 0.14 s & 2.41 min & $0.05\times$ \\
    Ditto (RoBERTa) & 2.64 s & 43.97 min & $1.00\times$ \\
    Qwen3-0.6B & 1.53 min & 25.43 h & $34.70\times$ \\
    Qwen3-1.7B & 1.57 min & 26.09 h & $35.60\times$ \\
    Qwen3-8B & 2.07 min & 34.53 h & $47.12\times$ \\
    \midrule
    \multicolumn{4}{l}{\textit{Zero-shot API teachers, 10 workers}} \\
    GPT-5.2 & 1.49 min & 24.83 h & $33.89\times$ \\
    Qwen 3.6 Plus & 12.50 min & 208.30 h & $284.27\times$ \\
    Kimi K2.6 & 20.17 min & 336.19 h & $458.80\times$ \\
    \bottomrule
  \end{tabular*}
\end{table}

\begin{table}[!t]
    \centering
    \caption{Composition of benchmark and machine-labeled training sets. Headers give hard-pair thresholds $\tau_p/\tau_n$.}
    \label{tab:dataset-comparison}
    \fontsize{7.5pt}{8.5pt}\selectfont
    \setlength{\tabcolsep}{2.8pt}
    \renewcommand{\arraystretch}{0.95}
    \begin{tabular*}{\columnwidth}{@{\extracolsep{\fill}}lrrrr@{}}
    \toprule
    Metric & Bench. & SS & AL-ML & AL-Ditto \\
    \midrule
    \multicolumn{5}{l}{\textit{Products}} \\
    \multicolumn{3}{l}{\textbf{\textsc{Abt-Buy}}} & \multicolumn{2}{r}{$\tau_p/\tau_n=0.14/0.17$} \\
    Positive pairs & 616 & 815 & 862 & 1{,}114 \\
    Negative pairs & 5{,}127 & 5{,}185 & 5{,}138 & 4{,}886 \\
    Entity clusters & 965 & 992 & 982 & 993 \\
    Positive rate & 10.73\% & 13.58\% & 14.37\% & 18.57\% \\
    Hard positives & 168 & 216 & 230 & 299 \\
    Hard negatives & 1{,}283 & 1{,}039 & 1{,}847 & 1{,}249 \\
    \midrule
    \multicolumn{3}{l}{\textbf{\textsc{Walmart-Amazon}}} & \multicolumn{2}{r}{$\tau_p/\tau_n=0.32/0.35$} \\
    Positive pairs & 576 & 471 & 750 & 867 \\
    Negative pairs & 5{,}568 & 5{,}673 & 5{,}250 & 5{,}250 \\
    Entity clusters & 736 & 753 & 759 & 796 \\
    Positive rate & 9.38\% & 7.67\% & 12.50\% & 14.17\% \\
    Hard positives & 148 & 122 & 203 & 230 \\
    Hard negatives & 1{,}437 & 547 & 1{,}437 & 1{,}094 \\
    \midrule
    \multicolumn{3}{l}{\textbf{Dn7 Walmart-Amazon}} & \multicolumn{2}{r}{$\tau_p/\tau_n=0.35/0.27$} \\
    Positive pairs & 457 & 460 & 903 & 908 \\
    Negative pairs & 25{,}593 & 25{,}590 & 25{,}147 & 25{,}142 \\
    Known clusters & 763 & 655 & 751 & 763 \\
    Positive rate & 1.75\% & 1.77\% & 3.47\% & 3.49\% \\
    Hard positives & 116 & 138 & 261 & 263 \\
    Hard negatives & 6{,}568 & 6{,}738 & 11{,}815 & 9{,}761 \\
    \midrule
    \multicolumn{3}{l}{\textbf{\textsc{WDC Products}}} & \multicolumn{2}{r}{$\tau_p/\tau_n=0.10/0.17$} \\
    Positive pairs & 8{,}471 & 5{,}000 & 5{,}000 & 10{,}951 \\
    Negative pairs & 11{,}364 & 15{,}000 & 15{,}000 & 8{,}884 \\
    Entity clusters & 500 & 500 & 499 & 500 \\
    Positive rate & 42.71\% & 25.00\% & 25.00\% & 55.21\% \\
    Hard positives & 2{,}118 & 819 & 1{,}035 & 1{,}879 \\
    Hard negatives & 2{,}846 & 2{,}475 & 7{,}497 & 3{,}819 \\
    \midrule
    \multicolumn{3}{l}{\textbf{billiger.de}} & \multicolumn{2}{r}{$\tau_p/\tau_n=0.08/0.12$} \\
    Positive pairs & 1{,}462 & 1{,}508 & 2{,}239 & 3{,}508 \\
    Negative pairs & 4{,}435 & 4{,}389 & 3{,}658 & 2{,}389 \\
    Known clusters & 500 & 493 & 479 & 492 \\
    Positive rate & 24.79\% & 25.57\% & 37.97\% & 59.49\% \\
    Hard positives & 366 & 256 & 199 & 493 \\
    Hard negatives & 1{,}109 & 1{,}904 & 3{,}054 & 1{,}643 \\
    \midrule
    \multicolumn{5}{l}{\textit{Literature}} \\
    \multicolumn{3}{l}{\textbf{\textsc{DBLP-ACM}}} & \multicolumn{2}{r}{$\tau_p/\tau_n=0.68/0.16$} \\
    Positive pairs & 1{,}332 & 1{,}416 & 2{,}142 & 2{,}195 \\
    Negative pairs & 6{,}085 & 6{,}001 & 5{,}275 & 4{,}956 \\
    Entity clusters & 1{,}952 & 2{,}026 & 2{,}082 & 2{,}083 \\
    Positive rate & 17.96\% & 19.09\% & 28.88\% & 30.70\% \\
    Hard positives & 335 & 384 & 595 & 615 \\
    Hard negatives & 1{,}537 & 1{,}196 & 2{,}808 & 2{,}368 \\
    \midrule
    \multicolumn{3}{l}{\textbf{\textsc{DBLP-Scholar}}} & \multicolumn{2}{r}{$\tau_p/\tau_n=0.47/0.20$} \\
    Positive pairs & 3{,}207 & 3{,}464 & 3{,}735 & 3{,}440 \\
    Negative pairs & 14{,}016 & 13{,}759 & 13{,}488 & 13{,}760 \\
    Entity clusters & 2{,}211 & 2{,}197 & 2{,}273 & 2{,}294 \\
    Positive rate & 18.62\% & 20.11\% & 21.69\% & 20.00\% \\
    Hard positives & 802 & 946 & 1{,}325 & 969 \\
    Hard negatives & 3{,}695 & 2{,}242 & 4{,}644 & 4{,}815 \\
    \midrule
    \multicolumn{3}{l}{\textbf{Semi-HETER}} & \multicolumn{2}{r}{$\tau_p/\tau_n=0.44/0.42$} \\
    Positive pairs & 474 & 444 & 289 & 275 \\
    Negative pairs & 766 & 796 & 951 & 965 \\
    Known clusters & 482 & 424 & 322 & 352 \\
    Positive rate & 38.23\% & 35.81\% & 23.31\% & 22.18\% \\
    Hard positives & 120 & 105 & 91 & 79 \\
    Hard negatives & 196 & 58 & 360 & 281 \\
    \bottomrule
    \end{tabular*}
\end{table}

\paragraph{Repeated use.} Under the WDC Products token profile, one million direct teacher decisions would cost about \$506, whereas the recorded similarity-search acquisition cost \$13.54 before preprocessing and student training. A trained student avoids these recurring API calls. A cheaper direct matcher lowers this recurring cost but keeps it: our GPT-5-mini review runs cost \$17.45 for 56,059 pairs at Batch rates, about \$311 per million decisions, whereas a trained student costs only local inference time (Table~\ref{tab:inference-scaling}). Local training and inference still need hardware, whose cost varies by deployment.

\noindent\textbf{Finding.} A training set costs a few dollars of teacher calls once, whereas direct matching pays per pair and exceeds this cost after tens of thousands of pairs on WDC Products. Active learning adds local committee training. The student infers 34 to 459 times faster than direct matching in our setup, so distillation pays off for large-scale matching where the student suffices (Tables~\ref{tab:labeling-costs} and~\ref{tab:inference-scaling}).

\subsection{Analysis of the Training Set Composition}
\label{sec:composition}

We analyze how pair selection shapes the training sets. Table~\ref{tab:dataset-comparison} compares the benchmark training sets (Bench.) with the sets generated by similarity search (SS), AL-ML, and AL-Ditto with GPT-5.2 in entity coverage, class balance, and hard pairs. Total pair counts match within 5\% across construction methods.

\textbf{Entity coverage:} Coverage changes little across construction methods on most benchmarks. On Abt-Buy, Walmart-Amazon, and the DBLP tasks, all four sets hold 3.4 to 8.4 labeled pairs per entity cluster. Every training set of WDC Products and billiger.de covers about 500 entity clusters, so students see few distinct products. Coverage alone does not track student F1. AL-Ditto on Dn7 covers the same 763 known clusters as the benchmark set, and the Semi-HETER AL-ML student exceeds benchmark-trained F1 by 10.15 points with 322 instead of 482 known clusters (Table~\ref{tab:selection-strategies}).

\textbf{Class balance:} AL-Ditto raises the positive rate over benchmark training on seven benchmarks, from 1.4 points on DBLP-Scholar to 34.7 on billiger.de, and AL-ML raises it on six (Table~\ref{tab:dataset-comparison}). Similarity search changes it by less than 3 points, except on WDC Products, where it and AL-ML stay at 25\% positives and AL-Ditto at 55\%.

\textbf{Hard pairs:} We measure surface similarity $s$ as all-field token Jaccard over lower-cased \texttt{[a-z0-9]+} tokens from all shared non-reserved attributes. Based on the corner-case definition of \citet{peeters2024wdcproducts}, hard positives are label-1 pairs with low similarity ($s \le \tau_p$), and hard negatives are label-0 pairs with high similarity ($s \ge \tau_n$). Each benchmark has fixed cutoffs for all construction methods. Table~\ref{tab:dataset-comparison} lists the rounded $\tau_p/\tau_n$, and the artifact the exact values.

AL-Ditto contains more hard positives than benchmark training on six benchmarks, for example 615 versus 335 on DBLP-ACM, and fewer on WDC Products (1{,}879 versus 2{,}118) and Semi-HETER (79 versus 120). It contains more hard negatives on six benchmarks and fewer only on Abt-Buy and Walmart-Amazon. Counts and rates can diverge. On billiger.de, AL-Ditto raises hard positives from 366 to 493, but their share among positive pairs falls from 25.03\% to 14.05\%, and its student still trails benchmark training by 5.09 F1.

Semi-HETER, where the machine-labeled students gain most, reverses the pattern. Both active-learning sets lower the positive rate from 38.23\% to 23.31\% and 22.18\%, concentrate on 322 and 352 instead of 482 known clusters, and contain more hard negatives (360 and 281 versus 196). The similarity-search student also gains 8.35 F1 with 424 known clusters and only 58 hard negatives, so not all of the gain comes from active learning.

\noindent\textbf{Finding.} Pair selection changes class balance and hard-pair counts more than entity coverage. Neither broader coverage nor more hard pairs goes with higher student F1, and no composition property identifies the most useful training set across benchmarks (Table~\ref{tab:dataset-comparison}).

\subsection{Analysis of Label Errors}
\label{sec:error-analysis}

A human audit estimates the error rates of teacher predictions and benchmark gold labels on the test populations. It covers the teachers' zero-shot predictions on test pairs rather than the labels of the generated training sets. For Abt-Buy, Walmart-Amazon, WDC Products, DBLP-ACM, and DBLP-Scholar, we audit every test pair on which at least one of the three teachers disagrees with the gold label, plus 50 control pairs per benchmark on which all three agree with it (25 gold-positive and 25 gold-negative). In total, 1{,}028 test pairs are audited, 778 disagreements and 250 controls. The first author was the sole annotator. The interface shows both records and hides the gold label and teacher decisions. The annotator consulted web search when external evidence was needed and marked a pair as ambiguous when the evidence remained insufficient. The five ambiguous pairs are excluded, and the error rates are population estimates weighted by the inverse sampling probability (Table~\ref{tab:labeler-error-rate}).

\begin{table}[!ht]
  \centering
  \caption{Estimated label error rates from the weighted human audit. Lowest bold.}
  \label{tab:labeler-error-rate}
  \small
  \setlength{\tabcolsep}{4.5pt}
  \begin{tabular}{lrrrr}
    \toprule
    Benchmark & Gold & GPT-5.2 & Qwen 3.6 Plus & Kimi K2.6 \\
    \midrule
    \multicolumn{5}{l}{\textit{Products}} \\
    \textsc{Abt-Buy} & 1.71\% & 2.28\% & 2.34\% & \textbf{1.60\%} \\
    \textsc{Walmart-Amazon} & 1.42\% & 1.61\% & 2.20\% & \textbf{0.98\%} \\
    \textsc{WDC Products} & 1.91\% & 1.82\% & 2.36\% & \textbf{1.60\%} \\
    \midrule
    \multicolumn{5}{l}{\textit{Literature}} \\
    \textsc{DBLP-ACM} & \textbf{0.53\%} & 0.81\% & 1.82\% & 0.65\% \\
    \textsc{DBLP-Scholar} & 2.52\% & 3.13\% & 4.49\% & \textbf{2.28\%} \\
    \bottomrule
  \end{tabular}
\end{table}

The gold labels disagree with the audit on an estimated 0.53\% of the test pairs of DBLP-ACM up to 2.52\% of DBLP-Scholar, and the disagreements include clear gold-label errors. Two examples, on which the published label contradicts both the audit and all three teachers, are an Abt-Buy pair of Apple Time Capsules with different capacities and model numbers and a DBLP-Scholar pair with near-identical titles but different authors and venues. Both are labeled as matches and documented in the artifact. Error rates over all test pairs do not translate directly into F1 points, so the audit leaves open how much these errors affect small gaps between students, such as 1.79 F1 on DBLP-Scholar (Table~\ref{tab:selection-strategies}).

Teacher error rates are of similar quantity. All estimates in Table~\ref{tab:labeler-error-rate} fall between 0.53\% and 4.49\%, and the teacher estimates differ from the gold-label estimates by 0.5 points on average and by at most 2 points. Among the teachers, Kimi K2.6 has the lowest estimate on all five tasks, averaging 0.2 points below the gold labels, and Qwen 3.6 Plus the highest, averaging 1.0 point above.

\noindent\textbf{Finding.} On the five audited test sets, teacher predictions and benchmark labels have similar estimated error rates (Table~\ref{tab:labeler-error-rate}). Small F1 differences may therefore partly reflect gold-label errors.
\section{Related Work}
\label{sec:related}

We position our study within entity matching, PLM- and LLM-based matching, machine labels, active learning, and distillation.

\textbf{Entity Matching:}
Entity matching methods range from probabilistic record linkage~\cite{fellegi1969record} and rule-based systems to supervised feature-based and deep models~\cite{elmagarmid2007duplicate,christen2012data,mudgal2018deep,barlaug2021survey}. Surveys describe the wider data-integration process~\cite{christophides2020overview,christophides2015webdata}, and systems such as Magellan support data preparation, blocking, labeling, and evaluation~\cite{konda2016magellan}. Benchmark studies characterize how difficult a matching task is. Primpeli and Bizer~\cite{primpeli2020profiling} profile 21 benchmark tasks and estimate their difficulty with baseline matchers, Papadakis et al.~\cite{papadakis2024critical} find most established benchmarks easy for learning-based matchers and rebuild harder ones, and WDC Products controls the share of corner cases and unseen entities~\cite{peeters2024wdcproducts}. Our selection of the eight benchmarks builds on these studies (Section~\ref{sec:setup}).

\textbf{PLM-based Entity Matching:}
Neural matchers learn record representations that replace manually designed similarity functions~\cite{mudgal2018deep,barlaug2021survey}. A pretrained encoder can use the surrounding record context when comparing attribute values, rather than evaluating each string independently. Pretrained language model (PLM) matchers learn from serialized records and pairs~\cite{brunner2020transformer,li2020ditto}. Ditto combines pretrained encoders with matching-specific augmentation and domain knowledge~\cite{li2020ditto}. R-SupCon adds supervised contrastive pre-training on product offers~\cite{peeters2022supcon}, and HierGAT applies hierarchical graph attention over tokens and attributes~\cite{yao2022hiergat}. Their dependence on task-specific training data and difficulty with unseen entities~\cite{peeters2024wdcproducts,peeters2025llmem} motivate our analysis of machine-labeled sets.

\textbf{LLM-based Entity Matching:}
Prompted matching presents two serialized records to a generative model and requests a matching decision. This avoids fitting a task-specific classifier but requires an LLM call for each candidate pair. \citet{narayan2022foundation} report state-of-the-art entity matching results with a prompted foundation model, and \citet{peeters2025llmem} find interactions among prompt, model, and dataset in zero-shot and few-shot matching. To lower the cost per pair, BatchER groups several pairs and demonstrations into one prompt~\cite{fan2024batcher}. \citet{wang2025match} compare matching a pair, comparing candidates, and selecting the match from a candidate list, and find selection most effective. Fine-tuning studies report improvements for smaller models and mixed cross-domain transfer effects~\cite{steiner2025finetuning}. \citet{zhang2025deepdive} compare fine-tuned small models with prompted large models in cross-dataset matching and report comparable effectiveness at lower deployment cost, and \citet{zhang2026beyondscale} distinguish architecture, backbone, and size effects in heterogeneous matching.

\textbf{Learning from Machine-Generated Labels:}
Weak supervision combines the noisy votes of hand-written labeling functions into training labels~\cite{ratner2017snorkel}. LLMs also annotate other text tasks, with studies reporting quality comparable to crowd workers~\cite{gilardi2023chatgpt,li2023coannotating,bansal2023llmannotators} and task-dependent reliability~\cite{reiss2025chatbots}. Such labels contain errors, and label-noise research identifies likely errors from model predictions, for example with confident learning~\cite{northcutt2021confident}. Entity resolution adds a structural signal because match decisions should form consistent clusters, and clustering algorithms turn pairwise decisions into such clusters~\cite{hassanzadeh2009framework}. Our post-processing combines a second-model review of each label with a consistency check over the graph of positive labels (Section~\ref{sec:method}), and our audit compares teacher and benchmark labels with human judgments (Section~\ref{sec:error-analysis}).

\textbf{Active Learning for Entity Matching:}
Active learning allocates annotation effort to selected examples. For entity matching, a useful training set needs matches and difficult non-matches, whereas random candidate pairs are usually dominated by easy non-matches. The selection method therefore influences both annotation volume and the types of distinctions the student learns. ALIAS selects the record pairs on which a committee of classifiers disagrees most for interactive labeling~\cite{sarawagi2002interactive}. Later work combines active learning with transfer learning for deep matchers~\cite{kasai2019lowresource} and with jointly learned blocking for transformer matchers~\cite{jain2021dial}. Prior work also explores rule evolution~\cite{iselebizerVLDB2012}, bootstrapping, and graph-based multi-source matching~\cite{primpeli2020unsupervised,primpeli2021graph}. Under explicit budgets, Battleship concentrates annotation on candidate regions likely to contain matches~\cite{genossar2023battleship}, BEACON adds labels from other domains~\cite{pulsone2026beacon}, and ERAMD mines matching dependencies for high-confidence labels~\cite{zhang2025eramd}. Progressive matching prioritizes early matches under time constraints~\cite{maciejewski2025progressive}, while 3dSAGER learns blocking schemes from small labeled samples~\cite{genossar20253dsager}. Our comparison fixes LLM-based annotation and varies pair selection, including disagreement-based committees with feature models or Ditto.

\textbf{Knowledge Distillation for Entity Matching:}
Surveys describe transferring teacher outputs, rationales, or preferences to smaller models~\cite{xu2024survey,10.1145/3699518,fang2026knowledge}. DistillER~\cite{zeakis2026distiller} studies data selection, labels from one or several teachers, and supervised fine-tuning versus reinforcement learning with open-weight Llama and Qwen teachers. Its SELECT formulation labels a query record together with a candidate list, and its selection study compares ranking, clustering, and random sampling over blocking similarity.

Our study differs in its reference point. We train the same student on each machine-labeled set and on the released benchmark training set and evaluate both on the benchmark's test split. The generated sets replace both pairs and labels, so the comparison shows where a complete machine-labeled training set can take the place of manual annotation and how this depends on selection, teacher, and student. The evaluation also covers billiger.de, released after GPT-5.2's training cutoff, Semi-HETER with its structural variations, label post-processing, a human audit of teacher and benchmark labels, and student versus direct LLM cost. DistillER finds similarity ranking the most reliable unsupervised selection method. This agrees with our similarity-search results at large budgets and on the bibliographic tasks, whereas we find committee-based active learning often ahead at small budgets and on Dn7 at every budget, and we separate label quality from pair selection (Table~\ref{tab:fixed-pairs}). \citet{steiner2025finetuning} fine-tune LLMs on benchmark training sets and add generated explanations and training examples, whereas our workflows select and label the training pairs themselves. Explanations alongside labels yield moderate, model-dependent gains elsewhere~\cite{wadhwa2024learning}. Distillation also serves entity alignment~\cite{zhang2025distillation4entityalginment} and entity linking~\cite{liu-etal-2023-distillation4entitylinking}.
\section{Limitations}
\label{sec:limitations}

The older public benchmarks may have been part of the teachers' pretraining data. At least one teacher contradicts the published gold label on 778 audited test pairs, including pairs with wrong gold labels (Section~\ref{sec:error-analysis}). These disagreements argue against simple recall of the labels. The labels and splits of billiger.de were released after GPT-5.2's training cutoff (Section~\ref{sec:setup}), and GPT-5.2 still reaches 75.71 F1 zero-shot on this benchmark (Table~\ref{tab:llm-direct}). However, the exposure to the underlying product offers cannot be ruled out.

The reported standard deviations cover student training only. Each training set was acquired and labeled once, so the variability of pair selection and teacher labeling is not measured.

All teachers label with one general prompt. Prompt effects vary across models and datasets~\cite{peeters2025llmem}, so other prompts could change individual F1 values, while all comparisons share the prompt.

The eight benchmarks cover products, bibliographic records, and books, including German offers and records of differing structure. Established benchmarks from further domains, such as beer and restaurants, are easy for learning-based matchers~\cite{papadakis2024critical}, and Ditto reaches 1.0 F1 on the Machamp restaurant task~\cite{wang2021machamp}. Person and business matching, which relies mainly on names, dates, and addresses, remains open.

The source tables hold only the benchmark training records, which the benchmark authors selected and which are dense in matches. Larger, sparser tables remain untested, as does a broader product pool, which may narrow the gap on WDC Products and billiger.de.

The API costs are reconstructed from token logs at standard rates. Committee fit times cover recorded fits and exclude final student training, preprocessing, and idle allocation. Runtime comparisons reflect the reported hardware, batching, and API concurrency.
\section{Conclusion}
\label{sec:conclusion}
\balance
This study compared knowledge-distillation workflows for entity matching by varying pair selection, label budget, teacher, post-processing, and student, with the student trained on the benchmark training set as common reference. With GPT-5.2 as teacher, machine-labeled sets are a practical alternative to benchmark training for a PLM-based matcher on seven of eight benchmarks.

Pair selection matters most at small label budgets, where active learning with a PLM committee often needs fewer labels, and larger budgets narrow the differences on several tasks. The open-weight Kimi K2.6 trains competitive students. The effect of post-processing varies by task, which makes label cleaning an optional, task-specific step. A weighted human audit of five test sets finds teacher errors and benchmark label errors of similar quantity.

The compact student competes with larger students and stays close to its teacher on most benchmarks. On the two benchmarks with high shares of unseen products, the distilled and benchmark-trained student fall behind it, so label quality alone does not explain the gap.

For practitioners, the results suggest the following rules-of-thumb: use distillation instead of (or in addition to) manually labeling training data, select pairs using a PLM committee, label them with an LLM teacher, use open-weight models if records must stay local, treat label cleaning as optional, and use direct LLM matching instead of distillation for tasks involving many unseen entities.

Knowledge distillation with an LLM teacher is thus a practical alternative to labeling task-specific training sets in entity matching. A few dollars of teacher labels yield a student that matches benchmark training on most tasks and infers 34 to 459 times faster compared to directly using the LLM for matching.

\begin{acks}
The authors used Claude Code and OpenAI Codex to develop code and to revise the text. Figure 1 was generated using Codex image generation. They reviewed all AI-assisted content and take full responsibility for it. Study design, experiments, and the interpretation of the results are the authors' own.
\end{acks}

\bibliographystyle{ACM-Reference-Format}
\bibliography{references}


\begin{thebibliography}{52}


\ifx \showCODEN    \undefined \def \showCODEN     #1{\unskip}     \fi
\ifx \showDOI      \undefined \def \showDOI       #1{#1}\fi
\ifx \showISBNx    \undefined \def \showISBNx     #1{\unskip}     \fi
\ifx \showISBNxiii \undefined \def \showISBNxiii  #1{\unskip}     \fi
\ifx \showISSN     \undefined \def \showISSN      #1{\unskip}     \fi
\ifx \showLCCN     \undefined \def \showLCCN      #1{\unskip}     \fi
\ifx \shownote     \undefined \def \shownote      #1{#1}          \fi
\ifx \showarticletitle \undefined \def \showarticletitle #1{#1}   \fi
\ifx \showURL      \undefined \def \showURL       {\relax}        \fi
\providecommand\bibfield[2]{#2}
\providecommand\bibinfo[2]{#2}
\providecommand\natexlab[1]{#1}
\providecommand\showeprint[2][]{arXiv:#2}

\bibitem[\protect\citeauthoryear{Bansal and Sharma}{Bansal and Sharma}{2023}]%
        {bansal2023llmannotators}
\bibfield{author}{\bibinfo{person}{Parikshit Bansal} {and}
  \bibinfo{person}{Amit Sharma}.} \bibinfo{year}{2023}\natexlab{}.
\newblock \bibinfo{title}{Large Language Models as Annotators: Enhancing
  Generalization of {NLP} Models at Minimal Cost}.
\newblock
\newblock
\showeprint[arxiv]{2306.15766}~[cs.CL]


\bibitem[\protect\citeauthoryear{Barlaug and Gulla}{Barlaug and Gulla}{2021}]%
        {barlaug2021survey}
\bibfield{author}{\bibinfo{person}{Nils Barlaug} {and}
  \bibinfo{person}{Jon~Atle Gulla}.} \bibinfo{year}{2021}\natexlab{}.
\newblock \showarticletitle{Neural Networks for Entity Matching: A Survey}.
\newblock \bibinfo{journal}{\emph{ACM Transactions on Knowledge Discovery from
  Data}} \bibinfo{volume}{15}, \bibinfo{number}{3} (\bibinfo{year}{2021}),
  \bibinfo{pages}{52:1--52:37}.
\newblock


\bibitem[\protect\citeauthoryear{Brunner and Stockinger}{Brunner and
  Stockinger}{2020}]%
        {brunner2020transformer}
\bibfield{author}{\bibinfo{person}{Ursin Brunner} {and} \bibinfo{person}{Kurt
  Stockinger}.} \bibinfo{year}{2020}\natexlab{}.
\newblock \showarticletitle{Entity Matching with Transformer Architectures - A
  Step Forward in Data Integration}. In \bibinfo{booktitle}{\emph{Proceedings
  of the 23rd International Conference on Extending Database Technology}}.
  \bibinfo{publisher}{OpenProceedings.org}, \bibinfo{address}{Konstanz,
  Germany}, \bibinfo{pages}{463--473}.
\newblock


\bibitem[\protect\citeauthoryear{Christen}{Christen}{2012}]%
        {christen2012data}
\bibfield{author}{\bibinfo{person}{Peter Christen}.}
  \bibinfo{year}{2012}\natexlab{}.
\newblock \bibinfo{booktitle}{\emph{Data Matching: Concepts and Techniques for
  Record Linkage, Entity Resolution, and Duplicate Detection}}.
\newblock \bibinfo{publisher}{Springer}, \bibinfo{address}{Berlin, Heidelberg}.
\newblock


\bibitem[\protect\citeauthoryear{Christophides, Efthymiou, Palpanas, Papadakis,
  and Stefanidis}{Christophides et~al\mbox{.}}{2020}]%
        {christophides2020overview}
\bibfield{author}{\bibinfo{person}{Vassilis Christophides},
  \bibinfo{person}{Vasilis Efthymiou}, \bibinfo{person}{Themis Palpanas},
  \bibinfo{person}{George Papadakis}, {and} \bibinfo{person}{Kostas
  Stefanidis}.} \bibinfo{year}{2020}\natexlab{}.
\newblock \showarticletitle{An Overview of End-to-End Entity Resolution for Big
  Data}.
\newblock \bibinfo{journal}{\emph{Comput. Surveys}} \bibinfo{volume}{53},
  \bibinfo{number}{6} (\bibinfo{year}{2020}), \bibinfo{pages}{127:1--127:42}.
\newblock


\bibitem[\protect\citeauthoryear{Christophides, Efthymiou, and
  Stefanidis}{Christophides et~al\mbox{.}}{2015}]%
        {christophides2015webdata}
\bibfield{author}{\bibinfo{person}{Vassilis Christophides},
  \bibinfo{person}{Vasilis Efthymiou}, {and} \bibinfo{person}{Kostas
  Stefanidis}.} \bibinfo{year}{2015}\natexlab{}.
\newblock \bibinfo{booktitle}{\emph{Entity Resolution in the Web of Data}}.
\newblock \bibinfo{publisher}{Morgan \& Claypool}, \bibinfo{address}{San
  Rafael, CA, USA}.
\newblock


\bibitem[\protect\citeauthoryear{Elmagarmid, Ipeirotis, and
  Verykios}{Elmagarmid et~al\mbox{.}}{2007}]%
        {elmagarmid2007duplicate}
\bibfield{author}{\bibinfo{person}{Ahmed~K. Elmagarmid},
  \bibinfo{person}{Panagiotis~G. Ipeirotis}, {and}
  \bibinfo{person}{Vassilios~S. Verykios}.} \bibinfo{year}{2007}\natexlab{}.
\newblock \showarticletitle{Duplicate Record Detection: A Survey}.
\newblock \bibinfo{journal}{\emph{IEEE Transactions on Knowledge and Data
  Engineering}} \bibinfo{volume}{19}, \bibinfo{number}{1}
  (\bibinfo{year}{2007}), \bibinfo{pages}{1--16}.
\newblock


\bibitem[\protect\citeauthoryear{Fan, Han, Fan, Chai, Tang, et~al\mbox{.}}{Fan
  et~al\mbox{.}}{2024}]%
        {fan2024batcher}
\bibfield{author}{\bibinfo{person}{Meihao Fan}, \bibinfo{person}{Xiaoyue Han},
  \bibinfo{person}{Ju Fan}, \bibinfo{person}{Chengliang Chai},
  \bibinfo{person}{Nan Tang}, {et~al\mbox{.}}} \bibinfo{year}{2024}\natexlab{}.
\newblock \showarticletitle{Cost-Effective In-Context Learning for Entity
  Resolution: A Design Space Exploration}. In \bibinfo{booktitle}{\emph{2024
  IEEE 40th International Conference on Data Engineering}}
  \emph{(\bibinfo{series}{ICDE 2024})}. \bibinfo{publisher}{IEEE Computer
  Society}, \bibinfo{address}{Los Alamitos, CA, USA},
  \bibinfo{pages}{3696--3709}.
\newblock


\bibitem[\protect\citeauthoryear{Fang, Yu, Cai, Chen, Wu, et~al\mbox{.}}{Fang
  et~al\mbox{.}}{2026}]%
        {fang2026knowledge}
\bibfield{author}{\bibinfo{person}{Luyang Fang}, \bibinfo{person}{Xiaowei Yu},
  \bibinfo{person}{Jiazhang Cai}, \bibinfo{person}{Yongkai Chen},
  \bibinfo{person}{Shushan Wu}, {et~al\mbox{.}}}
  \bibinfo{year}{2026}\natexlab{}.
\newblock \showarticletitle{Knowledge distillation and dataset distillation of
  large language models: Emerging trends, challenges, and future directions}.
\newblock \bibinfo{journal}{\emph{Artificial Intelligence Review}}
  \bibinfo{volume}{59}, \bibinfo{number}{1} (\bibinfo{year}{2026}),
  \bibinfo{pages}{17}.
\newblock


\bibitem[\protect\citeauthoryear{Fellegi and Sunter}{Fellegi and
  Sunter}{1969}]%
        {fellegi1969record}
\bibfield{author}{\bibinfo{person}{Ivan~P. Fellegi} {and}
  \bibinfo{person}{Alan~B. Sunter}.} \bibinfo{year}{1969}\natexlab{}.
\newblock \showarticletitle{A Theory for Record Linkage}.
\newblock \bibinfo{journal}{\emph{J. Amer. Statist. Assoc.}}
  \bibinfo{volume}{64}, \bibinfo{number}{328} (\bibinfo{year}{1969}),
  \bibinfo{pages}{1183--1210}.
\newblock


\bibitem[\protect\citeauthoryear{Genossar, Dalyot, Shraga, and Gal}{Genossar
  et~al\mbox{.}}{2025}]%
        {genossar20253dsager}
\bibfield{author}{\bibinfo{person}{Bar Genossar}, \bibinfo{person}{Sagi
  Dalyot}, \bibinfo{person}{Roee Shraga}, {and} \bibinfo{person}{Avigdor Gal}.}
  \bibinfo{year}{2025}\natexlab{}.
\newblock \showarticletitle{{3dSAGER}: Geospatial Entity Resolution over {3D}
  Objects}.
\newblock \bibinfo{journal}{\emph{Proceedings of the ACM on Management of
  Data}} \bibinfo{volume}{3}, \bibinfo{number}{6} (\bibinfo{year}{2025}),
  \bibinfo{pages}{286:1--286:28}.
\newblock


\bibitem[\protect\citeauthoryear{Genossar, Gal, and Shraga}{Genossar
  et~al\mbox{.}}{2023}]%
        {genossar2023battleship}
\bibfield{author}{\bibinfo{person}{Bar Genossar}, \bibinfo{person}{Avigdor
  Gal}, {and} \bibinfo{person}{Roee Shraga}.} \bibinfo{year}{2023}\natexlab{}.
\newblock \showarticletitle{The Battleship Approach to the Low Resource Entity
  Matching Problem}.
\newblock \bibinfo{journal}{\emph{Proceedings of the ACM on Management of
  Data}} \bibinfo{volume}{1}, \bibinfo{number}{4} (\bibinfo{year}{2023}),
  \bibinfo{pages}{224:1--224:25}.
\newblock


\bibitem[\protect\citeauthoryear{Gilardi, Alizadeh, and Kubli}{Gilardi
  et~al\mbox{.}}{2023}]%
        {gilardi2023chatgpt}
\bibfield{author}{\bibinfo{person}{Fabrizio Gilardi}, \bibinfo{person}{Meysam
  Alizadeh}, {and} \bibinfo{person}{Ma{\"e}l Kubli}.}
  \bibinfo{year}{2023}\natexlab{}.
\newblock \showarticletitle{{ChatGPT} Outperforms Crowd Workers for
  Text-Annotation Tasks}.
\newblock \bibinfo{journal}{\emph{Proceedings of the National Academy of
  Sciences}} \bibinfo{volume}{120}, \bibinfo{number}{30}
  (\bibinfo{year}{2023}), \bibinfo{pages}{e2305016120}.
\newblock


\bibitem[\protect\citeauthoryear{Hassanzadeh, Chiang, Lee, and
  Miller}{Hassanzadeh et~al\mbox{.}}{2009}]%
        {hassanzadeh2009framework}
\bibfield{author}{\bibinfo{person}{Oktie Hassanzadeh}, \bibinfo{person}{Fei
  Chiang}, \bibinfo{person}{Hyun~Chul Lee}, {and} \bibinfo{person}{Ren{\'e}e~J.
  Miller}.} \bibinfo{year}{2009}\natexlab{}.
\newblock \showarticletitle{Framework for Evaluating Clustering Algorithms in
  Duplicate Detection}.
\newblock \bibinfo{journal}{\emph{Proceedings of the VLDB Endowment}}
  \bibinfo{volume}{2}, \bibinfo{number}{1} (\bibinfo{year}{2009}),
  \bibinfo{pages}{1282--1293}.
\newblock


\bibitem[\protect\citeauthoryear{Isele and Bizer}{Isele and Bizer}{2012}]%
        {iselebizerVLDB2012}
\bibfield{author}{\bibinfo{person}{Robert Isele} {and}
  \bibinfo{person}{Christian Bizer}.} \bibinfo{year}{2012}\natexlab{}.
\newblock \showarticletitle{Learning Expressive Linkage Rules using Genetic
  Programming}.
\newblock \bibinfo{journal}{\emph{Proceedings of the VLDB Endowment}}
  \bibinfo{volume}{5}, \bibinfo{number}{11} (\bibinfo{year}{2012}),
  \bibinfo{pages}{1638--1649}.
\newblock


\bibitem[\protect\citeauthoryear{Jain, Sarawagi, and Sen}{Jain
  et~al\mbox{.}}{2021}]%
        {jain2021dial}
\bibfield{author}{\bibinfo{person}{Arjit Jain}, \bibinfo{person}{Sunita
  Sarawagi}, {and} \bibinfo{person}{Prithviraj Sen}.}
  \bibinfo{year}{2021}\natexlab{}.
\newblock \showarticletitle{Deep Indexed Active Learning for Matching
  Heterogeneous Entity Representations}.
\newblock \bibinfo{journal}{\emph{Proceedings of the VLDB Endowment}}
  \bibinfo{volume}{15}, \bibinfo{number}{1} (\bibinfo{year}{2021}),
  \bibinfo{pages}{31--45}.
\newblock


\bibitem[\protect\citeauthoryear{Kasai, Qian, Gurajada, Li, and Popa}{Kasai
  et~al\mbox{.}}{2019}]%
        {kasai2019lowresource}
\bibfield{author}{\bibinfo{person}{Jungo Kasai}, \bibinfo{person}{Kun Qian},
  \bibinfo{person}{Sairam Gurajada}, \bibinfo{person}{Yunyao Li}, {and}
  \bibinfo{person}{Lucian Popa}.} \bibinfo{year}{2019}\natexlab{}.
\newblock \showarticletitle{Low-resource Deep Entity Resolution with Transfer
  and Active Learning}. In \bibinfo{booktitle}{\emph{Proceedings of the 57th
  Annual Meeting of the Association for Computational Linguistics}}
  \emph{(\bibinfo{series}{ACL 2019})}. \bibinfo{publisher}{Association for
  Computational Linguistics}, \bibinfo{address}{Florence, Italy},
  \bibinfo{pages}{5851--5861}.
\newblock


\bibitem[\protect\citeauthoryear{Konda, Das, {Suganthan G. C.}, Doan, Ardalan,
  et~al\mbox{.}}{Konda et~al\mbox{.}}{2016}]%
        {konda2016magellan}
\bibfield{author}{\bibinfo{person}{Pradap Konda}, \bibinfo{person}{Sanjib Das},
  \bibinfo{person}{Paul {Suganthan G. C.}}, \bibinfo{person}{AnHai Doan},
  \bibinfo{person}{Adel Ardalan}, {et~al\mbox{.}}}
  \bibinfo{year}{2016}\natexlab{}.
\newblock \showarticletitle{Magellan: Toward Building Entity Matching
  Management Systems}.
\newblock \bibinfo{journal}{\emph{Proceedings of the VLDB Endowment}}
  \bibinfo{volume}{9}, \bibinfo{number}{12} (\bibinfo{year}{2016}),
  \bibinfo{pages}{1197--1208}.
\newblock


\bibitem[\protect\citeauthoryear{K{\"o}pcke, Thor, and Rahm}{K{\"o}pcke
  et~al\mbox{.}}{2010}]%
        {koepcke2010evaluation}
\bibfield{author}{\bibinfo{person}{Hanna K{\"o}pcke}, \bibinfo{person}{Andreas
  Thor}, {and} \bibinfo{person}{Erhard Rahm}.} \bibinfo{year}{2010}\natexlab{}.
\newblock \showarticletitle{Evaluation of Entity Resolution Approaches on
  Real-World Match Problems}.
\newblock \bibinfo{journal}{\emph{Proceedings of the VLDB Endowment}}
  \bibinfo{volume}{3}, \bibinfo{number}{1-2} (\bibinfo{year}{2010}),
  \bibinfo{pages}{484--493}.
\newblock


\bibitem[\protect\citeauthoryear{Kristensen-McLachlan, Canavan, K{\'a}rdos,
  Jacobsen, and Aar{\o}e}{Kristensen-McLachlan et~al\mbox{.}}{2025}]%
        {reiss2025chatbots}
\bibfield{author}{\bibinfo{person}{Ross~Deans Kristensen-McLachlan},
  \bibinfo{person}{Miceal Canavan}, \bibinfo{person}{Marton K{\'a}rdos},
  \bibinfo{person}{Mia Jacobsen}, {and} \bibinfo{person}{Lene Aar{\o}e}.}
  \bibinfo{year}{2025}\natexlab{}.
\newblock \showarticletitle{Are Chatbots Reliable Text Annotators? Sometimes}.
\newblock \bibinfo{journal}{\emph{PNAS Nexus}} \bibinfo{volume}{4},
  \bibinfo{number}{4} (\bibinfo{year}{2025}), \bibinfo{pages}{pgaf069}.
\newblock


\bibitem[\protect\citeauthoryear{Li, Shi, Ziems, Kan, Chen, et~al\mbox{.}}{Li
  et~al\mbox{.}}{2023}]%
        {li2023coannotating}
\bibfield{author}{\bibinfo{person}{Minzhi Li}, \bibinfo{person}{Taiwei Shi},
  \bibinfo{person}{Caleb Ziems}, \bibinfo{person}{Min-Yen Kan},
  \bibinfo{person}{Nancy Chen}, {et~al\mbox{.}}}
  \bibinfo{year}{2023}\natexlab{}.
\newblock \showarticletitle{{CoAnnotating}: Uncertainty-Guided Work Allocation
  between Human and Large Language Models for Data Annotation}. In
  \bibinfo{booktitle}{\emph{Proceedings of the 2023 Conference on Empirical
  Methods in Natural Language Processing}}. \bibinfo{publisher}{Association for
  Computational Linguistics}, \bibinfo{address}{Singapore},
  \bibinfo{pages}{1487--1505}.
\newblock


\bibitem[\protect\citeauthoryear{Li, Li, Suhara, Doan, and Tan}{Li
  et~al\mbox{.}}{2020}]%
        {li2020ditto}
\bibfield{author}{\bibinfo{person}{Yuliang Li}, \bibinfo{person}{Jinfeng Li},
  \bibinfo{person}{Yoshihiko Suhara}, \bibinfo{person}{AnHai Doan}, {and}
  \bibinfo{person}{Wang-Chiew Tan}.} \bibinfo{year}{2020}\natexlab{}.
\newblock \showarticletitle{Deep Entity Matching with Pre-Trained Language
  Models}.
\newblock \bibinfo{journal}{\emph{Proceedings of the VLDB Endowment}}
  \bibinfo{volume}{14}, \bibinfo{number}{1} (\bibinfo{year}{2020}),
  \bibinfo{pages}{50--60}.
\newblock


\bibitem[\protect\citeauthoryear{Liu, Tian, Lian, Wang, Cao, et~al\mbox{.}}{Liu
  et~al\mbox{.}}{2023}]%
        {liu-etal-2023-distillation4entitylinking}
\bibfield{author}{\bibinfo{person}{Yi Liu}, \bibinfo{person}{Yuan Tian},
  \bibinfo{person}{Jianxun Lian}, \bibinfo{person}{Xinlong Wang},
  \bibinfo{person}{Yanan Cao}, {et~al\mbox{.}}}
  \bibinfo{year}{2023}\natexlab{}.
\newblock \showarticletitle{Towards Better Entity Linking with Multi-View
  Enhanced Distillation}. In \bibinfo{booktitle}{\emph{Proceedings of the 61st
  Annual Meeting of the Association for Computational Linguistics (Volume 1:
  Long Papers)}}, \bibfield{editor}{\bibinfo{person}{Anna Rogers},
  \bibinfo{person}{Jordan Boyd-Graber}, {and} \bibinfo{person}{Naoaki Okazaki}}
  (Eds.). \bibinfo{publisher}{Association for Computational Linguistics},
  \bibinfo{address}{Toronto, Canada}, \bibinfo{pages}{9729--9743}.
\newblock


\bibitem[\protect\citeauthoryear{Maciejewski, Nikoletos, Papadakis, and
  Velegrakis}{Maciejewski et~al\mbox{.}}{2025}]%
        {maciejewski2025progressive}
\bibfield{author}{\bibinfo{person}{Jakub Maciejewski},
  \bibinfo{person}{Konstantinos Nikoletos}, \bibinfo{person}{George Papadakis},
  {and} \bibinfo{person}{Yannis Velegrakis}.} \bibinfo{year}{2025}\natexlab{}.
\newblock \showarticletitle{Progressive Entity Matching: A Design Space
  Exploration}.
\newblock \bibinfo{journal}{\emph{Proceedings of the ACM on Management of
  Data}} \bibinfo{volume}{3}, \bibinfo{number}{1} (\bibinfo{year}{2025}),
  \bibinfo{pages}{65:1--65:25}.
\newblock


\bibitem[\protect\citeauthoryear{Mohammed, Budach, Feuerpfeil, Ihde, Nathansen,
  et~al\mbox{.}}{Mohammed et~al\mbox{.}}{2025}]%
        {FelixDataQualityandML2025}
\bibfield{author}{\bibinfo{person}{Sedir Mohammed}, \bibinfo{person}{Lukas
  Budach}, \bibinfo{person}{Moritz Feuerpfeil}, \bibinfo{person}{Nina Ihde},
  \bibinfo{person}{Andrea Nathansen}, {et~al\mbox{.}}}
  \bibinfo{year}{2025}\natexlab{}.
\newblock \showarticletitle{The Effects of Data Quality on Machine Learning
  Performance on Tabular Data}.
\newblock \bibinfo{journal}{\emph{Information Systems}}  \bibinfo{volume}{132}
  (\bibinfo{year}{2025}), \bibinfo{pages}{102549}.
\newblock


\bibitem[\protect\citeauthoryear{Mudgal, Li, Rekatsinas, Doan, Park,
  et~al\mbox{.}}{Mudgal et~al\mbox{.}}{2018}]%
        {mudgal2018deep}
\bibfield{author}{\bibinfo{person}{Sidharth Mudgal}, \bibinfo{person}{Han Li},
  \bibinfo{person}{Theodoros Rekatsinas}, \bibinfo{person}{AnHai Doan},
  \bibinfo{person}{Youngchoon Park}, {et~al\mbox{.}}}
  \bibinfo{year}{2018}\natexlab{}.
\newblock \showarticletitle{Deep Learning for Entity Matching: A Design Space
  Exploration}. In \bibinfo{booktitle}{\emph{Proceedings of the 2018
  International Conference on Management of Data}}
  \emph{(\bibinfo{series}{SIGMOD '18})}. \bibinfo{publisher}{Association for
  Computing Machinery}, \bibinfo{address}{New York, NY, USA},
  \bibinfo{pages}{19--34}.
\newblock


\bibitem[\protect\citeauthoryear{Narayan, Chami, Orr, and R{\'e}}{Narayan
  et~al\mbox{.}}{2022}]%
        {narayan2022foundation}
\bibfield{author}{\bibinfo{person}{Avanika Narayan}, \bibinfo{person}{Ines
  Chami}, \bibinfo{person}{Laurel Orr}, {and} \bibinfo{person}{Christopher
  R{\'e}}.} \bibinfo{year}{2022}\natexlab{}.
\newblock \showarticletitle{Can Foundation Models Wrangle Your Data?}
\newblock \bibinfo{journal}{\emph{Proceedings of the VLDB Endowment}}
  \bibinfo{volume}{16}, \bibinfo{number}{4} (\bibinfo{year}{2022}),
  \bibinfo{pages}{738--746}.
\newblock


\bibitem[\protect\citeauthoryear{Northcutt, Jiang, and Chuang}{Northcutt
  et~al\mbox{.}}{2021}]%
        {northcutt2021confident}
\bibfield{author}{\bibinfo{person}{Curtis Northcutt}, \bibinfo{person}{Lu
  Jiang}, {and} \bibinfo{person}{Isaac Chuang}.}
  \bibinfo{year}{2021}\natexlab{}.
\newblock \showarticletitle{Confident Learning: Estimating Uncertainty in
  Dataset Labels}.
\newblock \bibinfo{journal}{\emph{Journal of Artificial Intelligence Research}}
   \bibinfo{volume}{70} (\bibinfo{year}{2021}), \bibinfo{pages}{1373--1411}.
\newblock


\bibitem[\protect\citeauthoryear{Papadakis, Kirielle, Christen, and
  Palpanas}{Papadakis et~al\mbox{.}}{2024}]%
        {papadakis2024critical}
\bibfield{author}{\bibinfo{person}{George Papadakis}, \bibinfo{person}{Nishadi
  Kirielle}, \bibinfo{person}{Peter Christen}, {and} \bibinfo{person}{Themis
  Palpanas}.} \bibinfo{year}{2024}\natexlab{}.
\newblock \showarticletitle{A Critical Re-evaluation of Record Linkage
  Benchmarks for Learning-Based Matching Algorithms}. In
  \bibinfo{booktitle}{\emph{2024 IEEE 40th International Conference on Data
  Engineering}} \emph{(\bibinfo{series}{ICDE 2024})}. \bibinfo{publisher}{IEEE
  Computer Society}, \bibinfo{address}{Los Alamitos, CA, USA},
  \bibinfo{pages}{3435--3448}.
\newblock


\bibitem[\protect\citeauthoryear{Peeters and Bizer}{Peeters and Bizer}{2022}]%
        {peeters2022supcon}
\bibfield{author}{\bibinfo{person}{Ralph Peeters} {and}
  \bibinfo{person}{Christian Bizer}.} \bibinfo{year}{2022}\natexlab{}.
\newblock \showarticletitle{Supervised Contrastive Learning for Product
  Matching}. In \bibinfo{booktitle}{\emph{Companion Proceedings of the Web
  Conference 2022}} \emph{(\bibinfo{series}{WWW '22})}.
  \bibinfo{publisher}{Association for Computing Machinery},
  \bibinfo{address}{New York, NY, USA}, \bibinfo{pages}{248--251}.
\newblock


\bibitem[\protect\citeauthoryear{Peeters, Der, and Bizer}{Peeters
  et~al\mbox{.}}{2024}]%
        {peeters2024wdcproducts}
\bibfield{author}{\bibinfo{person}{Ralph Peeters}, \bibinfo{person}{Reng~Chiz
  Der}, {and} \bibinfo{person}{Christian Bizer}.}
  \bibinfo{year}{2024}\natexlab{}.
\newblock \showarticletitle{{WDC} Products: A Multi-Dimensional Entity Matching
  Benchmark}. In \bibinfo{booktitle}{\emph{Proceedings of the 27th
  International Conference on Extending Database Technology}}
  \emph{(\bibinfo{series}{EDBT 2024})}.
  \bibinfo{publisher}{OpenProceedings.org}, \bibinfo{address}{Konstanz,
  Germany}, \bibinfo{pages}{22--33}.
\newblock


\bibitem[\protect\citeauthoryear{Peeters, Steiner, and Bizer}{Peeters
  et~al\mbox{.}}{2025}]%
        {peeters2025llmem}
\bibfield{author}{\bibinfo{person}{Ralph Peeters}, \bibinfo{person}{Aaron
  Steiner}, {and} \bibinfo{person}{Christian Bizer}.}
  \bibinfo{year}{2025}\natexlab{}.
\newblock \showarticletitle{Entity Matching using Large Language Models}. In
  \bibinfo{booktitle}{\emph{Proceedings of the 28th International Conference on
  Extending Database Technology}} \emph{(\bibinfo{series}{EDBT 2025})}.
  \bibinfo{publisher}{OpenProceedings.org}, \bibinfo{address}{Konstanz,
  Germany}, \bibinfo{pages}{529--541}.
\newblock


\bibitem[\protect\citeauthoryear{Primpeli and Bizer}{Primpeli and
  Bizer}{2020}]%
        {primpeli2020profiling}
\bibfield{author}{\bibinfo{person}{Anna Primpeli} {and}
  \bibinfo{person}{Christian Bizer}.} \bibinfo{year}{2020}\natexlab{}.
\newblock \showarticletitle{Profiling entity matching benchmark tasks}. In
  \bibinfo{booktitle}{\emph{Proceedings of the 29th ACM International
  Conference on Information \& Knowledge Management}}.
  \bibinfo{publisher}{Association for Computing Machinery},
  \bibinfo{address}{New York, NY, USA}, \bibinfo{pages}{3101--3108}.
\newblock


\bibitem[\protect\citeauthoryear{Primpeli and Bizer}{Primpeli and
  Bizer}{2021}]%
        {primpeli2021graph}
\bibfield{author}{\bibinfo{person}{Anna Primpeli} {and}
  \bibinfo{person}{Christian Bizer}.} \bibinfo{year}{2021}\natexlab{}.
\newblock \showarticletitle{Graph-boosted active learning for multi-source
  entity resolution}. In \bibinfo{booktitle}{\emph{International Semantic Web
  Conference}}. \bibinfo{publisher}{Springer}, \bibinfo{address}{Cham},
  \bibinfo{pages}{182--199}.
\newblock


\bibitem[\protect\citeauthoryear{Primpeli, Bizer, and Keuper}{Primpeli
  et~al\mbox{.}}{2020}]%
        {primpeli2020unsupervised}
\bibfield{author}{\bibinfo{person}{Anna Primpeli}, \bibinfo{person}{Christian
  Bizer}, {and} \bibinfo{person}{Margret Keuper}.}
  \bibinfo{year}{2020}\natexlab{}.
\newblock \showarticletitle{Unsupervised bootstrapping of active learning for
  entity resolution}. In \bibinfo{booktitle}{\emph{European Semantic Web
  Conference}}. \bibinfo{publisher}{Springer}, \bibinfo{address}{Cham},
  \bibinfo{pages}{215--231}.
\newblock


\bibitem[\protect\citeauthoryear{Pulsone, Shraga, and Goren}{Pulsone
  et~al\mbox{.}}{2026}]%
        {pulsone2026beacon}
\bibfield{author}{\bibinfo{person}{Nicholas Pulsone}, \bibinfo{person}{Roee
  Shraga}, {and} \bibinfo{person}{Gregory Goren}.}
  \bibinfo{year}{2026}\natexlab{}.
\newblock \showarticletitle{{BEACON}: Budget-Aware Entity Matching Across
  Domains}.
\newblock \bibinfo{journal}{\emph{Proceedings of the ACM on Management of
  Data}} \bibinfo{volume}{4}, \bibinfo{number}{3} (\bibinfo{year}{2026}),
  \bibinfo{pages}{144:1--144:26}.
\newblock


\bibitem[\protect\citeauthoryear{Ratner, Bach, Ehrenberg, Fries, Wu,
  et~al\mbox{.}}{Ratner et~al\mbox{.}}{2017}]%
        {ratner2017snorkel}
\bibfield{author}{\bibinfo{person}{Alexander Ratner},
  \bibinfo{person}{Stephen~H. Bach}, \bibinfo{person}{Henry Ehrenberg},
  \bibinfo{person}{Jason Fries}, \bibinfo{person}{Sen Wu}, {et~al\mbox{.}}}
  \bibinfo{year}{2017}\natexlab{}.
\newblock \showarticletitle{Snorkel: Rapid Training Data Creation with Weak
  Supervision}.
\newblock \bibinfo{journal}{\emph{Proceedings of the VLDB Endowment}}
  \bibinfo{volume}{11}, \bibinfo{number}{3} (\bibinfo{year}{2017}),
  \bibinfo{pages}{269--282}.
\newblock


\bibitem[\protect\citeauthoryear{Sarawagi and Bhamidipaty}{Sarawagi and
  Bhamidipaty}{2002}]%
        {sarawagi2002interactive}
\bibfield{author}{\bibinfo{person}{Sunita Sarawagi} {and}
  \bibinfo{person}{Anuradha Bhamidipaty}.} \bibinfo{year}{2002}\natexlab{}.
\newblock \showarticletitle{Interactive Deduplication Using Active Learning}.
  In \bibinfo{booktitle}{\emph{Proceedings of the Eighth ACM SIGKDD
  International Conference on Knowledge Discovery and Data Mining}}
  \emph{(\bibinfo{series}{KDD '02})}. \bibinfo{publisher}{Association for
  Computing Machinery}, \bibinfo{address}{New York, NY, USA},
  \bibinfo{pages}{269--278}.
\newblock


\bibitem[\protect\citeauthoryear{Steiner, Elagin, Peeters, Knopp, and
  Bizer}{Steiner et~al\mbox{.}}{2026}]%
        {billiger2026benchmark}
\bibfield{author}{\bibinfo{person}{Aaron Steiner}, \bibinfo{person}{Ksenia
  Elagin}, \bibinfo{person}{Ralph Peeters}, \bibinfo{person}{Johannes Knopp},
  {and} \bibinfo{person}{Christian Bizer}.} \bibinfo{year}{2026}\natexlab{}.
\newblock \bibinfo{title}{{Billiger.de} Products: A Bilingual Entity Matching
  Benchmark}.
\newblock
\newblock
\showeprint[arxiv]{2609.37713}~[cs.DB]


\bibitem[\protect\citeauthoryear{Steiner, Peeters, and Bizer}{Steiner
  et~al\mbox{.}}{2025}]%
        {steiner2025finetuning}
\bibfield{author}{\bibinfo{person}{Aaron Steiner}, \bibinfo{person}{Ralph
  Peeters}, {and} \bibinfo{person}{Christian Bizer}.}
  \bibinfo{year}{2025}\natexlab{}.
\newblock \showarticletitle{Fine-Tuning Large Language Models for Entity
  Matching}. In \bibinfo{booktitle}{\emph{2025 IEEE 41st International
  Conference on Data Engineering Workshops}} \emph{(\bibinfo{series}{ICDEW
  2025})}. \bibinfo{publisher}{IEEE Computer Society}, \bibinfo{address}{Los
  Alamitos, CA, USA}, \bibinfo{pages}{9--17}.
\newblock


\bibitem[\protect\citeauthoryear{Wadhwa, Krishnan, Wang, Wallace, and
  Kong}{Wadhwa et~al\mbox{.}}{2024}]%
        {wadhwa2024learning}
\bibfield{author}{\bibinfo{person}{Somin Wadhwa}, \bibinfo{person}{Adit
  Krishnan}, \bibinfo{person}{Runhui Wang}, \bibinfo{person}{Byron~C. Wallace},
  {and} \bibinfo{person}{Luyang Kong}.} \bibinfo{year}{2024}\natexlab{}.
\newblock \showarticletitle{Learning from Natural Language Explanations for
  Generalizable Entity Matching}. In \bibinfo{booktitle}{\emph{Proceedings of
  the 2024 Conference on Empirical Methods in Natural Language Processing}}.
  \bibinfo{publisher}{Association for Computational Linguistics},
  \bibinfo{address}{Miami, FL, USA}, \bibinfo{pages}{6114--6129}.
\newblock


\bibitem[\protect\citeauthoryear{Wang, Li, and Hirota}{Wang
  et~al\mbox{.}}{2021}]%
        {wang2021machamp}
\bibfield{author}{\bibinfo{person}{Jin Wang}, \bibinfo{person}{Yuliang Li},
  {and} \bibinfo{person}{Wataru Hirota}.} \bibinfo{year}{2021}\natexlab{}.
\newblock \showarticletitle{Machamp: A Generalized Entity Matching Benchmark}.
  In \bibinfo{booktitle}{\emph{Proceedings of the 30th ACM International
  Conference on Information \& Knowledge Management}}
  \emph{(\bibinfo{series}{CIKM '21})}. \bibinfo{publisher}{Association for
  Computing Machinery}, \bibinfo{address}{New York, NY, USA},
  \bibinfo{pages}{4633--4642}.
\newblock


\bibitem[\protect\citeauthoryear{Wang, Chen, Lin, Chen, Han,
  et~al\mbox{.}}{Wang et~al\mbox{.}}{2025}]%
        {wang2025match}
\bibfield{author}{\bibinfo{person}{Tianshu Wang}, \bibinfo{person}{Xiaoyang
  Chen}, \bibinfo{person}{Hongyu Lin}, \bibinfo{person}{Xuanang Chen},
  \bibinfo{person}{Xianpei Han}, {et~al\mbox{.}}}
  \bibinfo{year}{2025}\natexlab{}.
\newblock \showarticletitle{Match, Compare, or Select? {A}n Investigation of
  Large Language Models for Entity Matching}. In
  \bibinfo{booktitle}{\emph{Proceedings of the 31st International Conference on
  Computational Linguistics}} \emph{(\bibinfo{series}{COLING 2025})}.
  \bibinfo{publisher}{Association for Computational Linguistics},
  \bibinfo{address}{Abu Dhabi, UAE}, \bibinfo{pages}{96--109}.
\newblock


\bibitem[\protect\citeauthoryear{Wang, Lin, Fu, Han, Sun, et~al\mbox{.}}{Wang
  et~al\mbox{.}}{2022}]%
        {wang2022bridging}
\bibfield{author}{\bibinfo{person}{Tianshu Wang}, \bibinfo{person}{Hongyu Lin},
  \bibinfo{person}{Cheng Fu}, \bibinfo{person}{Xianpei Han},
  \bibinfo{person}{Le Sun}, {et~al\mbox{.}}} \bibinfo{year}{2022}\natexlab{}.
\newblock \showarticletitle{Bridging the Gap between Reality and Ideality of
  Entity Matching: A Revisiting and Benchmark Re-Construction}. In
  \bibinfo{booktitle}{\emph{Proceedings of the Thirty-First International Joint
  Conference on Artificial Intelligence}} \emph{(\bibinfo{series}{IJCAI
  2022})}. \bibinfo{publisher}{ijcai.org}, \bibinfo{address}{Vienna, Austria},
  \bibinfo{pages}{3978--3984}.
\newblock


\bibitem[\protect\citeauthoryear{Xu, Li, Tao, Shen, Cheng, et~al\mbox{.}}{Xu
  et~al\mbox{.}}{2024}]%
        {xu2024survey}
\bibfield{author}{\bibinfo{person}{Xiaohan Xu}, \bibinfo{person}{Ming Li},
  \bibinfo{person}{Chongyang Tao}, \bibinfo{person}{Tao Shen},
  \bibinfo{person}{Reynold Cheng}, {et~al\mbox{.}}}
  \bibinfo{year}{2024}\natexlab{}.
\newblock \bibinfo{title}{A Survey on Knowledge Distillation of Large Language
  Models}.
\newblock
\newblock
\showeprint[arxiv]{2402.13116}~[cs.CL]


\bibitem[\protect\citeauthoryear{Yang, Zhu, Lu, Wang, Chen, et~al\mbox{.}}{Yang
  et~al\mbox{.}}{2025}]%
        {10.1145/3699518}
\bibfield{author}{\bibinfo{person}{Chuanpeng Yang}, \bibinfo{person}{Yao Zhu},
  \bibinfo{person}{Wang Lu}, \bibinfo{person}{Yidong Wang},
  \bibinfo{person}{Qian Chen}, {et~al\mbox{.}}}
  \bibinfo{year}{2025}\natexlab{}.
\newblock \showarticletitle{Survey on Knowledge Distillation for Large Language
  Models: Methods, Evaluation, and Application}.
\newblock \bibinfo{journal}{\emph{ACM Transactions on Intelligent Systems and
  Technology}} \bibinfo{volume}{16}, \bibinfo{number}{6}
  (\bibinfo{year}{2025}), \bibinfo{pages}{1--27}.
\newblock


\bibitem[\protect\citeauthoryear{Yao, Gu, Cong, Jin, and Lv}{Yao
  et~al\mbox{.}}{2022}]%
        {yao2022hiergat}
\bibfield{author}{\bibinfo{person}{Dezhong Yao}, \bibinfo{person}{Yuhong Gu},
  \bibinfo{person}{Gao Cong}, \bibinfo{person}{Hai Jin}, {and}
  \bibinfo{person}{Xinqiao Lv}.} \bibinfo{year}{2022}\natexlab{}.
\newblock \showarticletitle{Entity Resolution with Hierarchical Graph Attention
  Networks}. In \bibinfo{booktitle}{\emph{Proceedings of the 2022 International
  Conference on Management of Data}} \emph{(\bibinfo{series}{SIGMOD '22})}.
  \bibinfo{publisher}{Association for Computing Machinery},
  \bibinfo{address}{New York, NY, USA}, \bibinfo{pages}{429--442}.
\newblock


\bibitem[\protect\citeauthoryear{Zeakis, Papadakis, Skoutas, and
  Koubarakis}{Zeakis et~al\mbox{.}}{2026}]%
        {zeakis2026distiller}
\bibfield{author}{\bibinfo{person}{Alexandros Zeakis}, \bibinfo{person}{George
  Papadakis}, \bibinfo{person}{Dimitrios Skoutas}, {and}
  \bibinfo{person}{Manolis Koubarakis}.} \bibinfo{year}{2026}\natexlab{}.
\newblock \bibinfo{title}{{DistillER}: Knowledge Distillation in Entity
  Resolution with Large Language Models}.
\newblock
\newblock
\showeprint[arxiv]{2602.05452}~[cs.DB]


\bibitem[\protect\citeauthoryear{Zhang, Wang, Huang, and Zhu}{Zhang
  et~al\mbox{.}}{2025c}]%
        {zhang2025eramd}
\bibfield{author}{\bibinfo{person}{Rong Zhang}, \bibinfo{person}{Dong Wang},
  \bibinfo{person}{Jintao Huang}, {and} \bibinfo{person}{Cheng Zhu}.}
  \bibinfo{year}{2025}\natexlab{c}.
\newblock \showarticletitle{Reliable Low-Resource Entity Resolution Enhanced by
  Rules}. In \bibinfo{booktitle}{\emph{2025 International Joint Conference on
  Neural Networks}} \emph{(\bibinfo{series}{IJCNN 2025})}.
  \bibinfo{publisher}{IEEE}, \bibinfo{address}{Piscataway, NJ, USA},
  \bibinfo{pages}{1--9}.
\newblock


\bibitem[\protect\citeauthoryear{Zhang, Song, Zhu, Bu, and Yu}{Zhang
  et~al\mbox{.}}{2025b}]%
        {zhang2025distillation4entityalginment}
\bibfield{author}{\bibinfo{person}{Yuhong Zhang}, \bibinfo{person}{Hangchi
  Song}, \bibinfo{person}{Xiaolong Zhu}, \bibinfo{person}{Chenyang Bu}, {and}
  \bibinfo{person}{Kui Yu}.} \bibinfo{year}{2025}\natexlab{b}.
\newblock \showarticletitle{An Robust Entity Alignment Method based on
  Knowledge Distillation with Noisy Aligned Pairs}. In
  \bibinfo{booktitle}{\emph{Proceedings of the 34th ACM International
  Conference on Information and Knowledge Management}} (Seoul, Republic of
  Korea) \emph{(\bibinfo{series}{CIKM '25})}. \bibinfo{publisher}{Association
  for Computing Machinery}, \bibinfo{address}{New York, NY, USA},
  \bibinfo{pages}{5510–5514}.
\newblock
\showISBNx{9798400720406}


\bibitem[\protect\citeauthoryear{Zhang, Groth, Calixto, and Schelter}{Zhang
  et~al\mbox{.}}{2025a}]%
        {zhang2025deepdive}
\bibfield{author}{\bibinfo{person}{Zeyu Zhang}, \bibinfo{person}{Paul Groth},
  \bibinfo{person}{Iacer Calixto}, {and} \bibinfo{person}{Sebastian Schelter}.}
  \bibinfo{year}{2025}\natexlab{a}.
\newblock \showarticletitle{A Deep Dive Into Cross-Dataset Entity Matching with
  Large and Small Language Models}. In \bibinfo{booktitle}{\emph{Proceedings of
  the 28th International Conference on Extending Database Technology}}
  \emph{(\bibinfo{series}{EDBT 2025})}.
  \bibinfo{publisher}{OpenProceedings.org}, \bibinfo{address}{Konstanz,
  Germany}, \bibinfo{pages}{922--934}.
\newblock


\bibitem[\protect\citeauthoryear{Zhang, Li, Calixto, Groth, and Schelter}{Zhang
  et~al\mbox{.}}{2026}]%
        {zhang2026beyondscale}
\bibfield{author}{\bibinfo{person}{Zeyu Zhang}, \bibinfo{person}{Xue Li},
  \bibinfo{person}{Iacer Calixto}, \bibinfo{person}{Paul Groth}, {and}
  \bibinfo{person}{Sebastian Schelter}.} \bibinfo{year}{2026}\natexlab{}.
\newblock \bibinfo{title}{Beyond Scale and Generation: Understanding Language
  Model-based Entity Matching}.
\newblock
\newblock
\showeprint[arxiv]{2607.24688}~[cs.DB]


\end{thebibliography}

\end{document}